\documentclass{article} 
\usepackage{iclr2027_conference,times}
\iclrfinalcopy

\usepackage{amsmath,amsfonts,bm}

\def\figref#1{figure~\ref{#1}}

\def\secref#1{section~\ref{#1}}

\def\eqref#1{equation~\ref{#1}}

\def\1{\bm{1}}

\DeclareMathAlphabet{\mathsfit}{\encodingdefault}{\sfdefault}{m}{sl}
\SetMathAlphabet{\mathsfit}{bold}{\encodingdefault}{\sfdefault}{bx}{n}

\usepackage{hyperref}
\usepackage{url}

\usepackage{microtype}
\usepackage{booktabs}
\usepackage{graphicx}
\usepackage{xspace}
\usepackage{subcaption}
\usepackage{amsmath}
\usepackage{wrapfig}
\usepackage{longtable}
\usepackage{booktabs}
\usepackage{enumitem}
\usepackage{multirow}
\usepackage{subcaption}
\usepackage{algorithm}
\usepackage{algorithmic}

\usepackage[export]{adjustbox}

\newcommand{\ours}{\textsc{RSD}\xspace}

\newcommand{\tabref}[1]{Table~\ref{#1}}

\newcommand{\appref}[1]{Appendix~\ref{#1}}
\renewcommand{\figref}[1]{Fig.~\ref{#1}}
\renewcommand{\secref}[1]{\S\ref{#1}}

\title{Self-Evolving Agents through Privileged Retrospective Distillation}

\author{
\begin{tabular}{l}
    Yuhang Zhou~~~~~~~
    Yifan Wu~~~~~~~
    Bo Peng~~~~~~~
    Mingyi Wang~~~~~~~
    Jiayi Liu~~~~~~~\\[0.3em]
    Xiangjun Fan~~~~~~~
    Lizhu Zhang~~~~~~
    Zhuokai Zhao~~~~~~
\end{tabular}\\\\
\begin{tabular}{l}
~~~Meta AI~~~~~\\[0.3em]
\texttt{~\{zyhang, zhuokai\}@meta.com}
\end{tabular}\\
}

\begin{document}

\maketitle
\begin{abstract}
    Self-evolving agents aim to improve from their own experience rather than remaining fixed once deployed.
In practice, however, naturally collected agent trajectories often lack ground-truth solutions or verifier-derived rewards, making them difficult to learn from.
We observe that these trajectories become substantially more informative in \textit{retrospect}. 
When the agent takes an action, it knows only what has happened so far, but looking back after the trajectory ends, the same action can be judged with knowledge the agent lacked at that moment, including how the environment responded afterward and how related tasks in its accumulated experience succeeded or failed.
Based on this observation, we introduce \textbf{Retrospective Self-Distillation (\ours)}, a multi-round self-evolution framework 
in which the same model, given privileged retrospective context built from its own experience, acts as a teacher that re-scores its own actions, and the resulting distribution is distilled back into the policy through on-policy self-distillation.
The privileged context combines cross-task precedents and lessons with turn-level reflection and future hindsight, allowing both successful and imperfect trajectories to provide dense supervision.
\ours is also built to make this supervision improve alongside the agent.
Every round, precedents and lessons are re-derived from the experience bank enlarged with the current policy's trajectories, and reflection and hindsight are computed fresh for each new trajectory, so the teacher always scores the policy against its most recent experience.
We further introduce \textit{frontier-guided context-policy evolution}, 
in which a stronger frontier model proposes edits to the context construction rules, for example how many precedents to retrieve or which components to include.
Across ALFWorld, ScienceWorld, and SearchQA with Qwen3-8B and 14B, \ours improves performance by up to $59.7\%$ relative over the base model, and frontier-guided context-policy evolution adds up to a further $44.7\%$ relative improvement over a fixed context policy.
These results show that agents can convert naturally accumulated experience into persistent model capability without ground-truth trajectories or verifier-derived rewards for the core policy update.
\end{abstract}

\section{Introduction}
\label{sec:intro}

Large language model (LLM) agents are increasingly capable of interacting with complex environments and completing multi-turn tasks.
Yet the substantial interaction experience an agent accumulates during use rarely makes it better at future tasks, and this has motivated growing interest in \textit{self-evolving agents}, which aim to establish a closed-loop \textit{data flywheel} where the current policy generates on-policy interaction experience, learns from that experience, and produces increasingly useful data in subsequent rounds~\citep{gao2025survey}.
Prior approaches pursue this goal along two lines.
One evolves non-parametric components such as prompts, workflows, memory, or tools~\citep{hu2025automated, hao2026self1}.
Experience improves future behavior in these systems, but the acquired capability is stored in the evolving scaffold rather than in the model, whose parameters remain fixed~\citep{lee2026recursive}.
The other line adapts model parameters using successful trajectories, supervised learning, or reinforcement learning~\citep{zhai2025agentevolver, meng2026rethinking}.
The improvement then persists in the model itself, but the update typically assumes supervision, such as verified successful trajectories or reliable rewards, that may not be naturally available during agent interaction.

In practical settings, each task may yield only a single on-policy trajectory, without a ground-truth trajectory specifying what the agent should have done or a reliable verifier providing rewards for policy optimization.
Conventional supervised fine-tuning therefore lacks direct target sequences, while verifier-based reinforcement learning requires reliable rewards and may be sample-inefficient when successful behavior must be discovered through additional exploration~\citep{wu2026seed}.
Restricting training to successful trajectories further discards much of the already collected experience, including failures that may contain informative environment feedback.
This raises a central question: \textit{How can an agent continuously improve from its accumulated interaction experience without requiring ground-truth trajectories or verifier-derived rewards for policy optimization?}

Our key observation is that although such trajectories may lack explicit training targets, they become substantially more informative \emph{after} the interaction has completed.
For the current task, downstream environment observations reveal the consequences of earlier decisions and provide hindsight unavailable when those actions were taken.
Across tasks, accumulated past experience provides concrete precedents, reusable strategies, and recurring failure patterns.
%

Based on this observation, we introduce \ours, a multi-round self-evolution framework that couples on-policy experience collection with privileged retrospective distillation.
At each round, the current policy generates new trajectories and incorporates them into an accumulated experience bank.
For every interaction turn, \ours then constructs a \emph{privileged retrospective} view of the agent's decision, combining \emph{precedents} retrieved from related past experience, \emph{lessons} distilled from accumulated successes and failures, \emph{reflection} on the current decision, and \emph{future hindsight} from downstream interaction and environment feedback.
This information asymmetry, between what the agent saw when acting and what is available afterward, naturally matches on-policy self-distillation (OPSD)~\citep{zhao2026self}: the same underlying model acts as both student and teacher, while the teacher additionally conditions on privileged information unavailable to the student during rollout.
Existing formulations construct this privileged view from ground-truth solutions~\citep{zhao2026self, xu2026beta}, whereas \ours draws it entirely from the agent's own evolving interaction experience.
The privileged teacher retrospectively re-scores the student's realized actions under this richer view, and the resulting supervision is distilled back into the policy.
As this process repeats, the model, its on-policy experience, and the privileged context constructed from that experience evolve together.
We further explore a stronger-access extension, \emph{frontier-guided context-policy evolution}, in which a frontier model also adapts how privileged context is constructed, yielding a self-evolution loop over both model capability and retrospective supervision.

We evaluate \ours with Qwen3-8B and Qwen3-14B~\citep{yang2025qwen3} across ALFWorld~\citep{shridhar2020alfworld}, ScienceWorld~\citep{wang2022scienceworld}, and SearchQA~\citep{jin2025search}.
\ours achieves up to a 59.7\% relative improvement over the base model, while frontier-guided context-policy evolution provides a further relative gain of up to $44.7\%$ over the fixed context policy.
We also find strong gains on ALFWorld, whereas SearchQA remains within a narrow performance band across rounds, providing a complementary boundary case for behavioral self-evolution.

Our main contributions are threefold:
\begin{itemize}[topsep=0pt,leftmargin=*,noitemsep]
    \item We formulate a practical self-evolving setting that turns naturally generated on-policy interaction experience into a recurrent data flywheel without requiring ground-truth trajectories or verifier-derived rewards for the core policy update.
    
    \item We introduce privileged retrospective distillation from accumulated experience and trajectory hindsight, and further extend it with frontier-guided context-policy evolution so that both the model and its retrospective supervision can evolve across rounds.
    
    \item Across ALFWorld, ScienceWorld, and SearchQA with Qwen3-8B and Qwen3-14B, \ours achieves up to $59.7\%$ relative improvement over the base model, while frontier-guided context-policy evolution yields up to an additional $44.7\%$ relative gain over the fixed context policy.
\end{itemize}
\section{Preliminaries}\label{subsec:preliminaries}

\subsection{Self-Evolving Agents}
\label{sec:prelim_self_evolution}

We consider a multi-turn agent that improves through a sequence of self-evolution rounds.
At round $r$, the agent is parameterized by a policy $\pi^{(r)}$.
Given a task $x$, the agent interacts with the environment over multiple turns.
At turn $i$, it observes the interaction history $h_i$ and generates an action
$
a_i \sim \pi^{(r)}(\cdot \mid h_i),
$
after which the environment returns a new observation $o_{i+1}$.
The resulting trajectory is denoted as
\[
\tau=(x,o_1,a_1,o_2,a_2,\ldots,o_{T+1}),
\qquad
\tau\sim\pi^{(r)}.
\]
Collecting trajectories over a set of tasks produces the round-$r$ on-policy experience
$\mathcal{D}^{(r)}$.

We focus on a practical self-evolution setting in which each sampled task produces a \textit{single} on-policy trajectory, rather than repeated attempts followed by success-based selection.
We further assume \textit{no ground-truth trajectories or verifier-derived rewards} for policy optimization, requiring the agent to extract useful supervision directly from its naturally collected interaction experience.

A self-evolving agent iteratively updates its policy using experience generated by its current policy,
\begin{equation}
\pi^{(r+1)}
=
\mathcal{U}\!\left(
\pi^{(r)},\mathcal{D}^{(r)}
\right),
\qquad
\mathcal{D}^{(r)}\sim\pi^{(r)},
\label{eq:self_evolution}
\end{equation}
where $\mathcal{U}$ denotes a generic self-improvement procedure.
Because $\pi^{(r+1)}$ induces a new interaction distribution, subsequent rounds continuously generate new on-policy experience for further improvement.
This iterative coupling between policy evolution and experience generation provides the basis for a self-evolving data flywheel.
In the following, we instantiate $\mathcal{U}$ using on-policy self-distillation.

\subsection{On-Policy Self-Distillation}
\label{sec:prelim_opsd}

On-policy self-distillation (OPSD) learns from trajectories generated by the current policy rather than from a fixed offline dataset.
The policy $\pi^{(r)}$ generates an on-policy trajectory
$\tau \sim \pi^{(r)}$.
OPSD uses the same underlying model to instantiate both student and teacher, with the teacher additionally receiving privileged context unavailable when the action was generated.
For a multi-turn agent, this privileged information is constructed at the turn level.
Let $c_i^{(r)}$ denote the privileged context for interaction turn $i$, and let $h_{i,j}$ denote the interaction history together with the generated action prefix preceding token $j$ at that turn.
The corresponding student and teacher distributions are
\[
\pi^{(r)}(\cdot \mid h_{i,j})
\qquad \text{and} \qquad
q^{(r)}(\cdot \mid h_{i,j},c_i^{(r)}),
\]
respectively, where $\pi^{(r)}$ and $q^{(r)}$ represent the unprivileged and privileged views of the same underlying model.
The student is then trained to match the privileged teacher distribution on its own on-policy states.
Aggregating token-level supervision across interaction turns gives
\begin{equation}
\mathcal{L}_{\mathrm{OPSD}}^{(r)}
=
\mathbb{E}_{\tau \sim \pi^{(r)}}\!
\left[
\sum_i \sum_j
D_{\mathrm{KL}}
\left(
q^{(r)}(\cdot \mid h_{i,j},c_i^{(r)})
\,\|\, 
\pi_{\theta}(\cdot \mid h_{i,j})
\right)
\right].
\label{eq:opsd}
\end{equation}
Here $\pi_\theta$ denotes the student under optimization, initialized from $\pi^{(r)}$. 
The sampled trajectories and the teacher distribution $q^{(r)}$ are held fixed during the update, and the updated student is taken as $\pi^{(r+1)}$ for the next round.
Because supervision is evaluated directly on the states visited by the current policy, OPSD can learn from imperfect and unsuccessful trajectories without requiring a ground-truth action sequence for each state.
\section{Retrospective Self-Distillation (\ours)}\label{sec:workflow}

\begin{figure}[t]
    \centering
    \includegraphics[width=\linewidth]{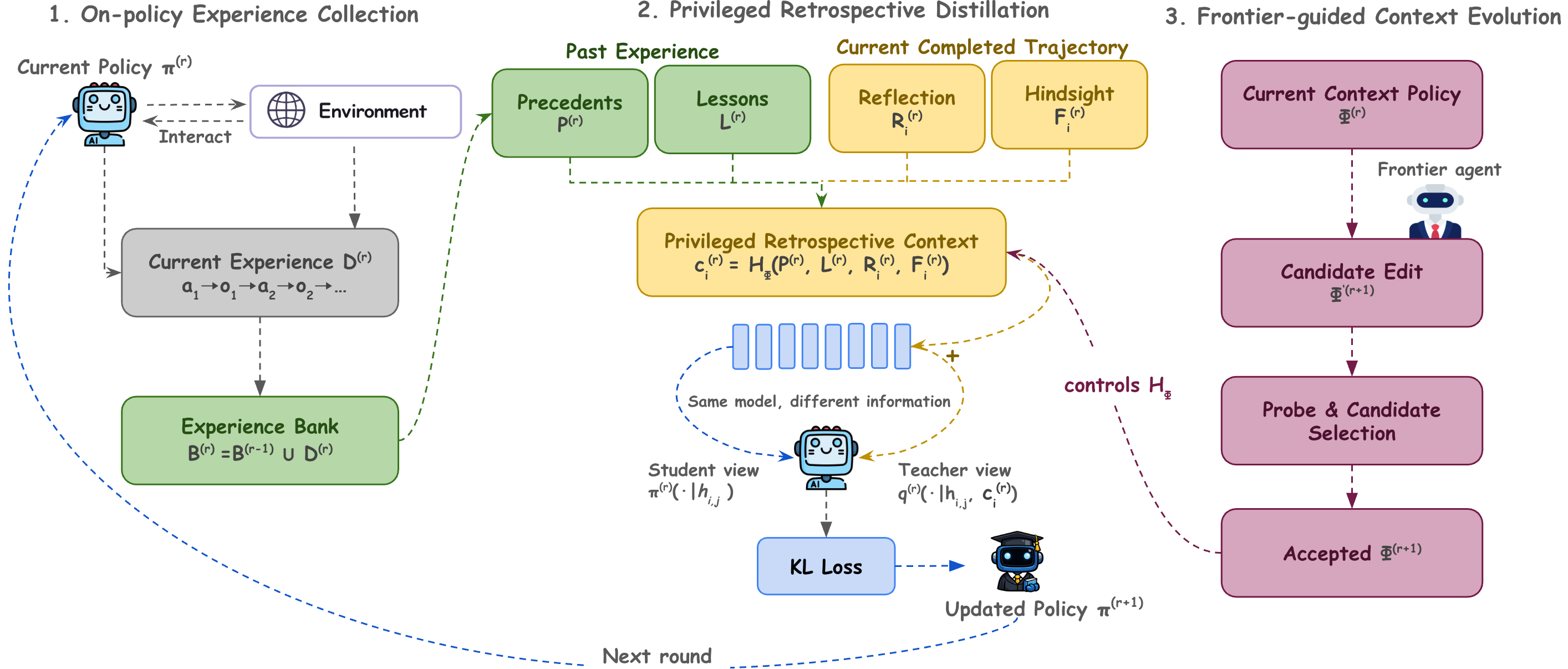}
    \caption{
    Overview of \ours.
    At self-evolution round $r$, the current policy $\pi^{(r)}$ first collects on-policy interaction experience, which is incorporated into the accumulated experience bank.
    Privileged retrospective context is then constructed from past experience: precedents and lessons, together with reflection and future hindsight from the completed current trajectory.
    The same underlying model retrospectively re-scores the student's realized actions under this privileged view, and the resulting distribution is distilled back into the student to obtain $\pi^{(r+1)}$.
    Repeating this process jointly evolves the policy, accumulated experience, and retrospective supervision.
    In the stronger-access extension, a frontier model additionally adapts the context-construction policy $\phi^{(r)}$ across rounds.
    }
    \label{fig:method_overview}
\end{figure}

Our goal is to build a self-evolving training pipeline in which an agent repeatedly generates on-policy interaction experience, learns from that experience, and uses the improved policy to produce increasingly useful data in subsequent rounds.
In the absence of ground-truth trajectories or reliable verifier-derived rewards, however, it is difficult to directly convert these rollouts into supervision through conventional supervised fine-tuning (SFT) or reward-based reinforcement learning (RLVR).
Nevertheless, the collected experience contains rich information that becomes available retrospectively: for the current task, the completed trajectory provides downstream environment feedback and interaction outcomes, while previous tasks contribute accumulated successful and failed experiences.
These signals provide a richer view of the agent's behavior than was available at action time.

This observation naturally motivates OPSD.
Rather than requiring an external target trajectory or scalar reward, we construct privileged retrospective context from the agent's accumulated experience and use it to re-score the agent's own on-policy trajectory.
The resulting privileged teacher distribution provides dense supervision on exactly the states visited by the current policy, which is then distilled back into the agent to produce the next generation.
As this process repeats, the evolving policy generates new experience that refreshes the privileged context, forming a data flywheel between model capability, interaction experience, and retrospective supervision.
Algorithm~\ref{alg:self_evolution} in \appref{sec:self_evolution_algorithm}
summarizes the complete agent self-evolving procedure.

\subsection{Self-Evolution through Privileged Retrospective Distillation}
\label{sec:self_evolution_method}

An overview of the complete self-evolution framework is shown in \figref{fig:method_overview}.
Our self-evolution pipeline alternates between on-policy experience collection and model update across rounds.
We use this round-based design so that each update is trained on trajectories sampled from a fixed current policy, after which the updated policy generates a refreshed on-policy distribution for the next round.
At round $r$, the current policy $\pi^{(r)}$ interacts with the environment and collects a new trajectory set $\mathcal{D}^{(r)}$.
We maintain an experience bank $\mathcal{B}^{(r)}$ containing all trajectories accumulated up to and including round $r$, including both successful and failed interactions:
\begin{equation}
\mathcal{B}^{(r)}
=
\mathcal{B}^{(r-1)} \cup \mathcal{D}^{(r)},
\qquad
\mathcal{B}^{(0)}=\emptyset.
\label{eq:experience_bank}
\end{equation}
Thus, $\mathcal{D}^{(r)}$ represents the newly collected on-policy experience at round $r$, while $\mathcal{B}^{(r)}$ provides the accumulated experience available for retrospective supervision in that round.

\paragraph{Privileged Retrospective Context.}
For each trajectory $\tau\in\mathcal{D}^{(r)}$ and interaction turn $i$, we construct turn-level privileged context
\begin{equation}
c_i^{(r)}
=
\mathcal{H}
\left(
P^{(r)},
L^{(r)},
R_i^{(r)},
F_i^{(r)}
\right),
\label{eq:privileged_context}
\end{equation}
where \textit{precedents} $P^{(r)}$ and \textit{lessons} $L^{(r)}$ are derived from the accumulated experience, while \textit{reflection} $R_i^{(r)}$ and \textit{future hindsight} $F_i^{(r)}$ exploit information available after completing the current trajectory.

\textit{Precedents} provide concrete prior experiences that demonstrate useful behaviors for solving related tasks.
We first apply task-specific heuristics to identify useful trajectories from $\mathcal{B}^{(r)}$, rank the resulting candidates by content-token overlap with the current task goal, and retrieve the top-$n$ as precedents (see~\appref{sec:implementation_details} for retrieval details).
These precedents complement the current trajectory with transferable action patterns observed in previous interactions.

\textit{Lessons} provide a more abstract view of accumulated experience.
For each task type, we expose selected successful and failed experiences of that type from $\mathcal{B}^{(r)}$ to the current policy model and ask it to summarize reusable strategies and recurring failure patterns.
The lesson summary is generated once per task type and round; we suppress the task-type index for notational simplicity:
$
L^{(r)}
=
\operatorname{Summarize}_{\pi^{(r)}}\!\left(\mathcal{B}^{(r)}\right).
$
Unlike precedents, lessons compress experience into reusable abstractions and allow unsuccessful trajectories to contribute even when they are unsuitable as concrete demonstrations.

The remaining two components are constructed retrospectively from the current trajectory.
At each turn $i$, the current policy model produces a \textit{reflection}
$
R_i^{(r)}
=
\operatorname{Reflect}_{\pi^{(r)}}(h_i,\tau),
$
which uses the completed interaction to diagnose the realized decision and summarize what could have been done more effectively.
We further provide \textit{future hindsight} $F_i^{(r)} = \tau_{>i},$ which directly exposes the downstream interaction following turn $i$, including subsequent environment feedback and outcome information.
%
%
Together, the four components provide the teacher with complementary cross-trajectory experience and within-trajectory hindsight that were unavailable to the acting policy at generation time.
The corresponding generation prompts are provided in \appref{sec:prompt_templates}.

\paragraph{Retrospective Distillation.}
Given $c_i^{(r)}$, we optimize the student using the OPSD objective in Eq.~\ref{eq:opsd}.
For every action token $j$ at turn $i$, the teacher conditions on $(h_{i,j},c_i^{(r)})$, whereas the student observes only $h_{i,j}$.
The teacher retrospectively re-scores the student's realized trajectory rather than generating a replacement trajectory, thereby providing dense supervision directly over the states visited by $\pi^{(r)}$.
After distillation, the updated student becomes $\pi^{(r+1)}$ and is used to collect $\mathcal{D}^{(r+1)}$.

\paragraph{Self-Evolution Flywheel.}
The interaction between policy evolution and experience accumulation forms a recurrent data flywheel.
At each round, $\pi^{(r)}$ generates new trajectories $\mathcal{D}^{(r)}$, which expand the experience bank to $\mathcal{B}^{(r)}$ through Eq.~\ref{eq:experience_bank} and refresh the precedents $P^{(r)}$ and lessons $L^{(r)}$ used for retrospective supervision.
Meanwhile, each newly collected trajectory provides fresh reflection $R_i^{(r)}$ and future hindsight $F_i^{(r)}$ for its own retrospective supervision.
The resulting privileged context in Eq.~\ref{eq:privileged_context} is distilled back into the policy, closing the flywheel between model evolution, accumulated experience, and privileged retrospective supervision.

\subsection{Frontier-Guided Context-Policy Evolution}
\label{sec:frontier_method}

The self-evolution procedure above continually refreshes the contents of privileged context as new experience enters the bank, while keeping the context-construction rule $\mathcal{H}$ fixed.
However, the most useful way to construct privileged supervision may itself change as the agent evolves.
We therefore consider a stronger-access setting that assumes access to both a frontier model and an executable verifier or environment metric, and uses them to optimize the context-construction policy.
Note that this external verifier is used only to evaluate context-policy candidates and their realized post-training performance; it does not provide rewards for optimizing the agent parameters, which continue to be updated solely through OPSD.

We parameterize the fixed privileged-context constructor $\mathcal{H}$ in Eq.~\ref{eq:privileged_context} by a context policy $\phi^{(r)}$, yielding $\mathcal{H}_{\phi^{(r)}}$.
The policy $\phi^{(r)}$ controls how the privileged information sources are selected, composed, and injected into the teacher context.
We represent $\phi^{(r)}$ as a declarative configuration over a constrained set of editable context choices, including precedent count and retrieval strategy, lesson usage, reflection, future hindsight, and the position at which privileged information is injected into the teacher context.
The frontier model may modify only these whitelisted choices, without modifying the model weights or training procedure.

\paragraph{Frontier-Guided Proposal.}
After completing evolution round $r$, the frontier model receives a compact summary of the current training state, including the current context policy, checkpoint task performance, and teacher re-scoring statistics such as correction rate, teacher entropy, and top-$k$ hit rate, with detailed definitions in \appref{sec:teacher_diagnostics}.
Based on these diagnostics, it proposes a small set of candidate edits to $\phi^{(r)}$.
To ground this search in empirical feedback, each candidate is evaluated with a lightweight no-training probe: we instantiate the candidate privileged context on a small training-distribution slice, provide it to the current model at inference time, and measure the corresponding task performance without updating model parameters.
The frontier model observes these probe scores and selects a candidate $\tilde{\phi}^{(r+1)}$ for the subsequent self-evolution round.

\paragraph{Retrospective Acceptance.}
The selected candidate $\tilde{\phi}^{(r+1)}$ is then used to construct privileged supervision during the subsequent full self-evolution round.
After training, we evaluate the resulting checkpoint using the actual environment metric.
The proposed context policy is retained if its post-training performance is no worse than the best previously accepted configuration (up to a tolerance $\epsilon$); otherwise, the context policy reverts to the last accepted configuration:
\begin{equation}
\phi^{(r+1)}
=
\begin{cases}
\tilde{\phi}^{(r+1)}, & m^{(r+1)} \geq m_{\mathrm{best}}^{(r)}-\epsilon,\\
\phi_{\mathrm{best}}^{(r)}, & \text{otherwise},
\end{cases}
\label{eq:context_acceptance}
\end{equation}
where $m^{(r+1)}$ denotes the realized post-training environment performance.
Only the context policy is reverted; the model parameters continue evolving across rounds.

\paragraph{Joint Evolution of Model and Context Policy.}
The frontier-guided extension therefore adds a second adaptive process to the self-evolution flywheel.
The core procedure evolves the model and the privileged information available from accumulated experience, while frontier guidance additionally evolves how that information is selected and organized:
\[
\pi^{(r)} \rightarrow \pi^{(r+1)},
\qquad
\mathcal{B}^{(r)} \rightarrow \mathcal{B}^{(r+1)},
\qquad
\phi^{(r)} \rightarrow \phi^{(r+1)}.
\]
This allows the retrospective supervision mechanism itself to adapt to the evolving agent rather than relying on a single hand-designed context policy throughout training.
\section{Experiments}\label{sec:experiments}

\subsection{Experimental Setup}
\label{sec:setup}

\paragraph{Models.}
We conduct experiments with Qwen3-8B and Qwen3-14B~\citep{yang2025qwen3} as the base agent models to study self-evolution across model scales.
For frontier-guided context-policy evolution, we use Claude-Opus-4.8~\citep{anthropic2026claudeopus48} as the frontier model that proposes constrained modifications to the privileged harness configuration.
We access Claude through OpenCode~\citep{opencode2026}, which serves as the coding and orchestration harness for generating and applying the proposed configuration edits.

\paragraph{Benchmarks and Evaluation Metrics.}
We evaluate \ours on three agent benchmarks: ALFWorld~\citep{shridhar2020alfworld}, ScienceWorld~\citep{wang2022scienceworld}, and SearchQA~\citep{jin2025search}.
Together, these benchmarks span embodied household interaction, scientific reasoning and environment manipulation, and search-based question answering, providing diverse settings for evaluating iterative agent self-evolution.
For ALFWorld, we report success rate (SR) on the valid seen and valid unseen splits.
For ScienceWorld, we report task success rate (SR) on the development set.
For SearchQA, we report exact match (EM).
Additional evaluation details are provided in \appref{sec:implementation_details}.

\paragraph{Baselines.}
We compare \ours against several representative adaptation and self-improvement baselines.
Base model (Base) directly evaluates the pretrained agent without additional adaptation, while Rejection Sampling Fine-Tuning (RFT)~\citep{yuan2023scaling} fine-tunes the model on its own successful trajectories.
Ground-Truth Supervised Fine-Tuning (GT-SFT) provides an oracle-supervision reference by fine-tuning on ground-truth trajectories, while Self-Taught Reasoner (STaR)~\citep{zelikman2022star} iteratively improves the model using self-generated successful trajectories.
We additionally consider BoN-16+SFT, which samples 16 candidate trajectories per training instance, selects successful trajectories using ground truth, and fine-tunes on the resulting data.
We further compare against Group Relative Policy Optimization (GRPO)~\citep{shao2024deepseekmath}, which performs verifier-based reinforcement learning when an executable reward is available, and Ground-Truth On-Policy Self-Distillation (GT-OPSD)~\citep{zhao2026self}, which replaces our privileged context with gold-reference agent trajectories during distillation.
For test-time adaptation, we include Reflexion@2~\citep{shinn2023reflexion}, which allows a second attempt conditioned on reflection from the preceding interaction, and Harness-at-Inference (Harness@Inf), which directly supplies the privileged context to the base model without post-training.

\paragraph{Implementation Details.}
We run self-evolution for 10 rounds, sampling 200 unique training instances in each round, and optimize the model using full-parameter forward-KL distillation with a learning rate of $3\times10^{-6}$.
Unless otherwise specified, \ours is evaluated using the evolved student alone without privileged context at inference time; additional training, decoding, and task-specific implementation details are provided in \appref{sec:implementation_details}.

\subsection{Main Results}
\label{sec:main_results}

\begin{table}[t]
\centering
\small
\setlength{\tabcolsep}{6pt}
\begin{tabular}{lcccccc}
\toprule
& \multicolumn{2}{c}{\textbf{ALFWorld (SR)}}
& \multicolumn{2}{c}{\textbf{ScienceWorld (SR)}}
& \multicolumn{2}{c}{\textbf{SearchQA (EM)}} \\
\cmidrule(lr){2-3}
\cmidrule(lr){4-5}
\cmidrule(lr){6-7}
\textbf{Method}
& \textbf{Qwen3-8B} & \textbf{Qwen3-14B}
& \textbf{Qwen3-8B} & \textbf{Qwen3-14B}
& \textbf{Qwen3-8B} & \textbf{Qwen3-14B} \\
\midrule

Base
& 62.9 & 69.3
& 15.6 & 12.9
& 35.4 & 35.4 \\

RFT
& 66.9 & 71.0
& 12.7 & 14.3
& 35.4 & 36.3 \\

GT-SFT
& 34.1 & 42.7
& 11.6 & 13.4
& 34.7 & 35.2 \\

STaR
& 65.3 & 77.8
& 12.7 & 13.6
& 34.4 & 37.6 \\

BoN-16+SFT
& 71.6 & 77.7
& \underline{20.8} & 20.4
& 29.7 & 35.5 \\

GRPO
& 75.5 & \textbf{82.6}
& -- & --
& 37.3 & 37.3 \\

GT-OPSD
& 70.1 & \underline{82.5}
& 3.0 & 3.0
& 37.1 & 38.5 \\

Reflexion@2
& \underline{76.0} & 76.7
& 17.8 & 18.5
& 37.1 & 38.1 \\

Harness@Inf
& 65.4 & 72.0
& 19.6 & \underline{20.5}
& \underline{37.7} & \textbf{41.4} \\

\textbf{\ours}
& \textbf{76.6} & 79.6
& \textbf{23.4} & \textbf{20.6}
& \textbf{37.9} & \underline{38.7} \\

\bottomrule
\end{tabular}
\caption{
Main results across ALFWorld, ScienceWorld, and SearchQA. For ALFWorld, SR is averaged over the valid-seen and valid-unseen splits, while for SearchQA, EM is averaged across the evaluated sub-datasets. Results are averaged over three seeds. \textbf{Bold} denotes the best result and \underline{underline} denotes the second-best result in each setting.
}
\label{tab:main_results}
\end{table}

\label{sec:main_results}

\tabref{tab:main_results} summarizes the main results across three benchmarks and two model scales.
Overall, \ours consistently improves over the corresponding base models and outperforms the evaluated self-training baselines across all six model--benchmark settings.
On ALFWorld-8B, \ours achieves an average SR of 76.6, compared with 62.9 for Base, 66.9 for RFT, 65.3 for STaR, and 71.6 for BoN-16+SFT.
The gains are particularly pronounced on ScienceWorld, where \ours improves from 15.6 to 23.4 at 8B and from 12.9 to 20.6 at 14B, while even BoN-16+SFT reaches only 20.8 and 20.4 despite substantially increasing rollout exploration.
These results suggest that increasing search for successful trajectories and subsequently fine-tuning on the selected experience does not by itself recover the benefit of privileged retrospective distillation.
In contrast, SearchQA exhibits a relatively narrow performance range across methods.
Our later analyses provide a consistent picture: performance remains largely flat across self-evolution rounds (\figref{fig:self_evolution_curves}), teacher corrections remain moderate but comparatively stable (\appref{sec:teacher_diagnostics}), and privileged context retains inference-time utility without being effectively internalized into the evolved model (\appref{sec:internalization}).
This suggests that the SearchQA boundary case reflects limited capacity to convert the available retrospective signal into persistent behavioral improvement, rather than simply an absence of useful contextual information.

\paragraph{Comparison with Reward-Driven and Oracle Baselines.}
\ours remains competitive with methods that rely on substantially stronger supervision.
On ALFWorld-14B, GRPO and GT-OPSD achieve 82.6 and 82.5 SR, respectively, compared with 79.6 for \ours.
However, GRPO relies on environment rewards and multiple grouped rollouts for policy optimization, substantially increasing rollout cost, while GT-OPSD explicitly conditions the teacher on a ground-truth successful agent trajectory.
In contrast, \ours learns from the agent's own interaction experience through privileged retrospective context, without requiring oracle demonstrations or reward-driven group optimization.
Notably, \ours achieves the best result on ALFWorld-8B and substantially outperforms the supervised baselines on ScienceWorld at both model scales.

\paragraph{Beyond Test-Time Adaptation.}
We further compare against Reflexion@2, which allows a second attempt conditioned on reflection from the preceding interaction, and Harness@Inf, which directly provides privileged context to the base model at inference time without updating model parameters.
Under the standard single-attempt evaluation protocol, \ours achieves higher point estimates than Reflexion@2 across all six model--benchmark settings and matches or exceeds Harness@Inf in five of six settings.
For example, on ScienceWorld-8B, Reflexion@2 and Harness@Inf achieve 17.8 and 19.6 SR, respectively, compared with 23.4 for \ours.
These results indicate that privileged retrospective supervision can be internalized into persistent model capability rather than remaining dependent on additional test-time adaptation.
A four-condition analysis comparing base and evolved models with and without privileged context further examines the extent of this internalization across tasks and model scales (\appref{sec:internalization}).
SearchQA-14B provides an important boundary case, where Harness@Inf reaches 41.4 compared with 38.7 for \ours, suggesting that some contextual information remains complementary to the model rather than being fully internalized.

Round-by-round trajectories show progressive self-evolution on ALFWorld and ScienceWorld, while SearchQA remains comparatively flat; full trajectories are provided in \appref{sec:self_evolution_analysis}.
\section{Analysis}\label{sec:generalization}

\subsection{Ablation Studies}
\label{sec:ablation}

\tabref{tab:context_ablation} studies the contribution of different components in the privileged context.
On ALFWorld, removing lessons, reflection, or future hindsight
each reduces SR by roughly five points, while removing precedents has only a minor effect.
\begin{wraptable}{r}{0.45\textwidth}
    \centering
    \small
    \setlength{\tabcolsep}{3pt}
    \begin{tabular}{lcc}
    \toprule
    \textbf{Variant} & \textbf{ALFWorld} & \textbf{ScienceWorld} \\
    \midrule
    Full              & 76.6 & 23.4 \\
    $-$ Precedents    & 75.0 & 15.1 \\
    $-$ Lessons & 71.6 & 22.7 \\
    $-$ Reflection    & 71.4 & 21.8 \\
    $-$ Future        & 71.4 & 17.6 \\
    \bottomrule
    \end{tabular}
    \caption{
    Ablation of privileged-context components on 8B.
    Different tasks rely on different privileged signals: generated hindsight components dominate on ALFWorld, whereas precedents are essential on ScienceWorld.
    }
    \label{tab:context_ablation}
\end{wraptable}
ScienceWorld exhibits a different pattern: removing precedents decreases SR from 23.4 to 15.1, nearly returning to the Base performance of 15.6, while removing future also causes a substantial drop to 17.6.
In comparison, lessons and reflection provide smaller gains on ScienceWorld.
To examine the source of the lesson signal, we construct lessons from successful or failed trajectories only.
Restricting lessons to either successful or failed trajectories yields lower performance than the default main-result reference, suggesting that both sources may provide useful retrospective information (\appref{sec:lesson_source_ablation}).
These results indicate that different environments benefit from different forms of privileged information, with trajectory-specific hindsight being more important on ALFWorld and transferable precedents being important on ScienceWorld.


\subsection{Effect of Context Evolution}

We next examine whether evolving the privileged context with newly collected experience is itself important for continued self-evolution (\figref{fig:context_evolution}).
Starting from the same base checkpoint, we compare the standard evolving-context configuration against variants that update only lessons, update only precedents, or freeze the round-1 context.
By round 3, the full configuration reaches 67.9 SR on ALFWorld, while updating lessons alone reaches 69.0, updating precedents alone reaches 63.8, and freezing the round-1 context drops to 59.7.
These results show that repeatedly training against stale retrospective context limits continued improvement, while refreshing the supervision with newly collected experience supports self-evolution.
Full evolution trajectories are provided in \appref{sec:context_evolution_diagnostic}.

\subsection{Frontier-Guided Context-Policy Evolution}
\label{sec:frontier_context}

\begin{figure}[t]
    \centering
    \includegraphics[width=0.9\linewidth]{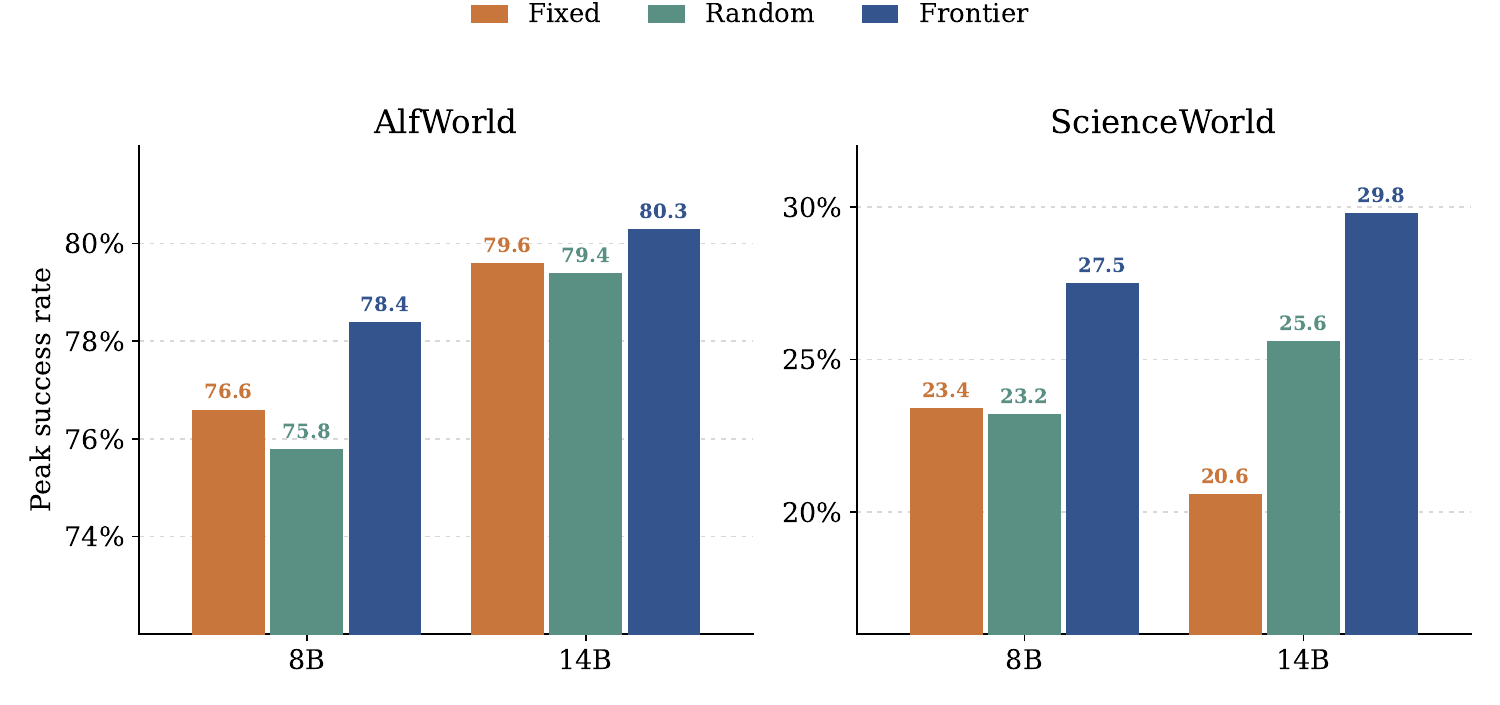}
    \caption{
        Performance of fixed, random, and frontier-guided context-policy evolution on ALFWorld and ScienceWorld.
        The frontier variant uses a frontier model to propose context-policy edits, with candidate selection guided by a no-training privileged-context-at-inference probe.
        We report success rate over self-evolution rounds for each task and model scale.
    }
    \label{fig:frontier_context_results}
\end{figure}

We next study a stronger-access extension in which privileged-context construction is itself optimized by a frontier model.
In this setting, we assume access to both a frontier model and an executable environment metric, which together enable an outer-loop search over context-construction policies.
At each round, the frontier model proposes a small YAML edit to the privileged-context policy from a constrained whitelist of editable knobs.
Candidate selection is performed using a cheap no-training privileged-context-at-inference probe, while retrospective acceptance or rejection is determined only after a full self-evolution round using the realized environment performance.
We compare this frontier-guided setting against two controls: \textit{Fixed}, which keeps the canonical privileged-context policy unchanged throughout training, and \textit{Random}, which explores the same edit space under the same edit budget but without frontier guidance.

\paragraph{Frontier Guidance Improves Context Evolution.}
As shown in \figref{fig:frontier_context_results}, frontier-guided context-policy evolution yields the clearest gains on ScienceWorld, where it improves SR from 23.4 to 27.5 at 8B and from 20.6 to 29.8 at 14B relative to the fixed policy, and also outperforms random search at both scales.
On ALFWorld, the gains are smaller but still positive, with SR improving from 76.6 to 78.4 at 8B and from 79.6 to 80.3 at 14B.
This pattern suggests that frontier-guided context optimization is most useful when privileged-context construction has substantial headroom, while remaining competitive in the more saturated ALFWorld regime.
We therefore focus the frontier-guided study on ALFWorld and ScienceWorld, where the preceding analyses reveal clearer self-evolution and context-construction headroom.
SearchQA serves instead as a complementary boundary case, where self-evolution remains limited and privileged context shows little evidence of persistent internalization into the model (\appref{sec:internalization}; \appref{sec:teacher_diagnostics}).
Ablations further show that the no-training probe is important for effective candidate selection; detailed probe construction and validation are provided in \appref{sec:frontier_probe}.
Across independent runs, the frontier also discovers systematic task-relevant context-policy adaptations; we analyze these retained edits and provide a representative evolution trajectory in \appref{sec:frontier_policy_analysis}.

\section{Conclusion}
\label{sec:conclusion}

We study self-evolving language agents that improve from their own on-policy interaction experience without requiring ground-truth trajectories or verifier-derived rewards for the core policy update.
We introduce \ours, which constructs privileged retrospective context from accumulated experience and trajectory hindsight, and distills this richer view back into the policy through on-policy self-distillation.
Across multiple agent benchmarks and model scales, \ours consistently improves over the corresponding base models.
Our ablation and context-evolution analyses further show that different privileged signals contribute complementary information and that continually refreshing the retrospective context is important for sustained self-evolution.
We additionally introduce frontier-guided context-policy evolution, allowing the construction of privileged supervision itself to adapt as the agent evolves.
Together, these results demonstrate a practical data flywheel in which model capability, accumulated experience, and retrospective supervision improve jointly across rounds.

\section*{AI use statement}
Generative AI tools were used for language polishing, proofreading, and assistance with coding during the preparation of this work. All AI-assisted text and code were reviewed, verified, and, where necessary, revised by the authors. We take responsibility for the final content of this work, including all text, claims, analyses, code, and other artifacts produced with the aid of generative AI.

\bibliography{iclr2027_conference}
\bibliographystyle{iclr2027_conference}

\appendix
\section*{Appendix}

\section{Related Work}
\label{sec:related}

\subsection{Self-Evolving Language Agents}
Self-evolving language agents aim to continuously improve from their own interaction experience rather than remaining fixed after post-training~\citep{gao2025survey, tao2024survey}.
One line of work primarily evolves the external agent system, including prompts, workflows, memory, tools, and other agentic structures~\citep{hu2025automated, zhang2025aflow, li2024autoflow, wu2026remember, hao2026self1}.
Automated agent-design methods search over workflow or program structures, while experience-driven agents continually refine explicit memories, reusable strategies, or dynamically constructed assets~\citep{yang2025learning, wu2025evolver, wei2025evo, cheng2026mem2evolve}.
These approaches demonstrate that accumulated interaction experience can improve future behavior, but much of the acquired capability remains mediated by an external scaffold rather than being persistently incorporated into the underlying policy~\citep{lee2026recursive}.

A complementary line of work directly adapts the agent model itself.
Recent approaches use online parameter updates, reinforcement learning, self-generated tasks, reward models, or hindsight supervision to internalize interaction experience into persistent model capability, reducing dependence on external scaffolds and enabling subsequent behavior to improve as new experience accumulates~\citep{zhai2025agentevolver, sun2025towards, ren2026scaling, wu2026seed, hao2026self2, meng2026rethinking}.
While these methods move self-evolution from the scaffold toward model weights, many rely on explicit verifiers, successful trajectories, or additional exploration to determine which experience should drive learning~\citep{zuo2026ttrl, zhang2026test}.
In contrast, we construct a data flywheel that does not use verifier-derived rewards for the core policy update, instead distilling privileged retrospective supervision from both the current trajectory and accumulated interactions directly into model weights, allowing the policy and its supervision to co-evolve across rounds.

\subsection{On-Policy Distillation and Privileged Context}
On-policy distillation (OPD) addresses the train--inference mismatch of conventional knowledge distillation by training the student on trajectories sampled from its own policy, while a teacher provides dense token-level supervision on those student-visited states~\citep{agarwal2024policy, yang2026learning, zhou2026omniopd}.
This formulation makes distillation naturally compatible with iterative policy improvement, since supervision is applied directly to the distribution induced by the evolving student.
Recent work extends this idea by changing what information is available to the teacher.
On-Policy Context Distillation (OPCD) conditions the teacher on additional context while the student generates without it, allowing useful in-context knowledge, such as historical solution traces or optimized prompts, to be internalized into model parameters~\citep{ye2026policy}.
On-Policy Self-Distillation (OPSD) further removes the need for a separate teacher model: the same underlying model acts as both teacher and student, with the teacher granted privileged information unavailable to the student; existing formulations commonly construct this privileged view from ground-truth solutions~\citep{zhao2026self, xu2026beta}.
Related self-distillation methods similarly exploit verified solutions, environment feedback, or successful rollouts as privileged signals~\citep{hubotter2026reinforcement,pan2026rlcsd,he2026self, hao2026self2}.

In contrast, our core method requires neither ground-truth trajectories nor verifier-derived rewards for model-parameter optimization.
We construct privileged retrospective context from the agent's own current trajectory and accumulated prior experience, and continually refresh this context as new on-policy experience is collected, coupling context evolution with model-weight evolution.

\section{Implementation Details}
\label{sec:implementation_details}

The prompt templates used for agent interaction, privileged-context
construction, and frontier-guided context-policy evolution are provided
in~\appref{sec:prompt_templates}.

\paragraph{Training Configuration.}
We perform self-evolution for 10 rounds and sample 200 training instances in each round.
Each round starts from the checkpoint produced by the previous round, while the accumulated experience bank is carried forward across rounds.
We use full-parameter fine-tuning with forward-KL top-$k$ distillation, where $k=20$, a learning rate of $3\times10^{-6}$, one training epoch per round, and a mini-batch size of 48.
No policy-gradient term or task-reward term is included in the OPSD objective.
All experiments are run on a single node with eight NVIDIA H200 GPUs.

\paragraph{Rollout Configuration.}
For ALFWorld, ScienceWorld, and SearchQA, the maximum numbers of interaction turns are 30, 15, and 15, respectively.
The maximum prompt/response lengths are $4096/8192$ tokens for ALFWorld, $2048/16384$ for ScienceWorld, and $4096/4096$ for SearchQA.
Training rollout temperatures are $1.0$ for ALFWorld, ScienceWorld, and SearchQA, respectively, with one rollout per training instance.
The larger rollout group used by GRPO is restricted to the GRPO baseline and is not used by \ours.

\paragraph{Teacher and Privileged-Context Configuration.}
The teacher retrospectively re-scores the student's on-policy trajectories under privileged context and provides the next-token distributions used for distillation.
By default, the privileged context consists of four components: precedents, accumulated lessons, turn-level reflection, and future hindsight over the current trajectory.
To construct the precedent set, we first heuristically filter candidate trajectories for successful or useful experiences.
For ALFWorld and ScienceWorld, success is determined by the environment completion signal, while for SearchQA we retain answers grounded in the top retrieved passages.
We then rank the retained precedents by content-token overlap with the current task goal and retrieve the top two by default ($n$=2). For precedent ranking, we compute content-token overlap between the current task goal and each candidate trajectory. We lowercase and tokenize the text, remove punctuation and non-content tokens such as stopwords, and rank candidates by the number of remaining tokens shared with the task goal.
Lessons accumulate reusable insights from previous experience, while reflection and future hindsight provide trajectory-specific retrospective information for the current turn.
Following the stabilization motivation of prior OPSD work~\citep{zhao2026self}, we prevent the teacher from evolving indefinitely.
The teacher initially tracks the evolving policy and is frozen at the best-so-far checkpoint once the benchmark selection metric decreases relative to its previous maximum; the frozen teacher is then used for all remaining rounds.
This choice is motivated by a controlled teacher-schedule ablation in \appref{sec:teacher_schedule_ablation}: a permanently frozen teacher remains stable but exhibits limited improvement, whereas a fully evolving self-teacher reaches substantially higher performance before collapsing in later rounds.
Unless otherwise specified, all reported results use these configurations.

\paragraph{Evaluation Protocol.}
For the main \ours evaluation, the evolved student is evaluated without
privileged context, such that the harness is used only to construct
training-time supervision.
Privileged context is provided at inference time only in explicitly labeled diagnostic settings, including Harness@Inf, the internalization analysis, and frontier candidate probes.
For ALFWorld, we evaluate approximately 140 valid-seen and 134 valid-unseen tasks and report success rate (SR) on both splits.
For ScienceWorld, we evaluate 449 development variations and report task success rate.
For SearchQA, we evaluate on the test set and report exact match (EM).
Inference uses vLLM with fixed decoding settings within each benchmark.

\paragraph{Frontier-Guided Context-Policy Evolution.}
The context policy is implemented as a YAML harness configuration and restricted to a predefined whitelist of context-policy and teacher-side re-scoring parameters, including precedent count and retrieval strategy, component activation, context position, and teacher temperature.
The frontier model is not permitted to modify the underlying training algorithm or context-construction code. 
For frontier-guided context-policy evolution, we set the retrospective acceptance tolerance to $\epsilon=0.5\%$, such that a proposed context policy is retained when its post-training environment performance is no more than $0.5$ percentage points below the best previously accepted configuration.

\paragraph{Baseline Implementations.}
We implement all comparison methods within the same training framework whenever possible and same training dataset size, changing only the optimization mode or supervision source.
GRPO disables distillation and performs group-relative policy optimization with a rollout group size of 8 using environment success rewards.
STaR iteratively fine-tunes the agent on its own successful trajectories.
GT-SFT uses oracle successful trajectories for supervised fine-tuning, whereas GT-OPSD retains the same on-policy distillation objective as \ours but conditions the teacher on a gold successful trajectory instead of the experience-derived privileged context.
For frontier-guided context-policy evolution, the Fixed Harness baseline keeps the canonical harness unchanged across rounds, while Random Harness samples edits from the same whitelisted YAML search space and under the same edit budget as the frontier-guided optimizer.

\section{Self-Evolution Algorithm}
\label{sec:self_evolution_algorithm}

Algorithm~\ref{alg:self_evolution} summarizes the complete training procedure.
The core method uses a fixed context-construction rule $\mathcal{H}$, while the
frontier-guided extension replaces it with the evolving policy
$\mathcal{H}_{\phi^{(r)}}$ described in \secref{sec:frontier_method}.

\begin{algorithm}[t]
\caption{\ours: Self-Evolution through Privileged Retrospective Distillation}
\label{alg:self_evolution}
\begin{algorithmic}[1]
\REQUIRE Initial policy $\pi^{(1)}$; task distribution $\mathcal{X}$;
number of rounds $R$; context-construction rule $\mathcal{H}$
\STATE Initialize experience bank $\mathcal{B}^{(0)} \leftarrow \emptyset$

\FOR{$r=1,\ldots,R$}

    \STATE Sample a set of tasks $\mathcal{X}^{(r)} \subset \mathcal{X}$

    \STATE Collect one on-policy trajectory for each task:
    \[
    \mathcal{D}^{(r)}
    \leftarrow
    \left\{
    \tau \sim \pi^{(r)}(\cdot \mid x)
    :
    x \in \mathcal{X}^{(r)}
    \right\}
    \]

    \STATE Update the experience bank:
    \[
    \mathcal{B}^{(r)}
    \leftarrow
    \mathcal{B}^{(r-1)} \cup \mathcal{D}^{(r)}
    \]

    \STATE Construct task-type lesson summaries from $\mathcal{B}^{(r)}$
    \[
    L^{(r)}
    \leftarrow
    \operatorname{Summarize}_{\pi^{(r)}}(\mathcal{B}^{(r)})
    \]

    \STATE Initialize $\mathcal{L}^{(r)} \leftarrow 0$

    \FOR{each trajectory $\tau \in \mathcal{D}^{(r)}$}

        \STATE Retrieve precedents $P^{(r)}$ from other tasks in $\mathcal{B}^{(r)}$

        \FOR{each interaction turn $i$ in $\tau$}

            \STATE Construct retrospective reflection:
            \[
            R_i^{(r)}
            \leftarrow
            \operatorname{Reflect}_{\pi^{(r)}}(h_i,\tau)
            \]

            \STATE Extract future hindsight:
            \[
            F_i^{(r)}
            \leftarrow
            \tau_{>i}
            \]

            \STATE Construct privileged retrospective context:
            \[
            c_i^{(r)}
            \leftarrow
            \mathcal{H}
            \left(
            P^{(r)},
            L^{(r)},
            R_i^{(r)},
            F_i^{(r)}
            \right)
            \]

            \FOR{each token $j$ in action $a_i$}

                \STATE Compute privileged teacher distribution:
                \[
                q^{(r)}_{i,j}
                \leftarrow
                q^{(r)}
                \left(
                \cdot \mid h_{i,j},c_i^{(r)}
                \right)
                \]

                \STATE Compute student distribution:
                \[
                p_{i,j}
                \leftarrow
                \pi_{\theta}
                \left(
                \cdot \mid h_{i,j}
                \right)
                \]

                \STATE Accumulate OPSD loss:
                \[
                \mathcal{L}^{(r)}
                \leftarrow
                \mathcal{L}^{(r)}
                +
                D_{\mathrm{KL}}
                \left(
                q^{(r)}_{i,j}
                \,\|\, p_{i,j}
                \right)
                \]

            \ENDFOR
        \ENDFOR
    \ENDFOR

    \STATE Update the student on its own on-policy states:
    \[
    \pi^{(r+1)}
    \leftarrow
    \arg\min_{\pi_{\theta}}
    \mathcal{L}^{(r)}
    \]

\ENDFOR

\RETURN $\pi^{(R+1)}$
\end{algorithmic}
\end{algorithm}

\paragraph{Frontier-Guided Extension.}
For frontier-guided context-policy evolution, we replace the fixed
$\mathcal{H}$ in Algorithm~\ref{alg:self_evolution} with
$\mathcal{H}_{\phi^{(r)}}$.
Between self-evolution rounds, the frontier model uses the current context policy,
training diagnostics, and no-training probe feedback to propose the next candidate
$\tilde{\phi}^{(r+1)}$.
After the corresponding full training round, the candidate is retained or reverted
according to Eq.~\ref{eq:context_acceptance}, with $\epsilon=0.5\%$.
Only the context policy may be reverted; the updated model parameters continue
to the next self-evolution round.

\section{Supplementary Results}
\label{sec:supplementary_results}

\subsection{Reflexion Test-Time Scaling}
\label{sec:reflexion_scaling}

To examine the effect of additional test-time interaction, we evaluate
Reflexion with one, two, and three sequential attempts.
Each subsequent attempt is conditioned on reflection from the preceding
interaction, allowing the agent to revise its behavior without updating
model parameters.
As shown in \tabref{tab:reflexion_scaling}, performance generally improves
as additional attempts are provided.
While \ours outperforms Reflexion@2 across all six settings under its
standard single-attempt evaluation, Reflexion@3 surpasses \ours in several
settings, illustrating that additional test-time interaction can trade
increased inference effort for higher performance.

\begin{table}[t]
\centering
\small
\setlength{\tabcolsep}{5pt}
\begin{tabular}{lcccccc}
\toprule
& \multicolumn{2}{c}{\textbf{ALFWorld (SR)}}
& \multicolumn{2}{c}{\textbf{ScienceWorld (SR)}}
& \multicolumn{2}{c}{\textbf{SearchQA (EM)}} \\
\cmidrule(lr){2-3}
\cmidrule(lr){4-5}
\cmidrule(lr){6-7}
\textbf{Method}
& \textbf{8B} & \textbf{14B}
& \textbf{8B} & \textbf{14B}
& \textbf{8B} & \textbf{14B} \\
\midrule
Reflexion@1
& 66.1 & 69.5
& 12.9 & 13.6
& 34.8 & 36.9 \\

Reflexion@2
& 76.0 & 76.7
& 17.8 & 18.5
& 37.1 & 38.1 \\

Reflexion@3
& 80.7 & 79.6
& 20.9 & 24.1
& 41.3 & 43.6 \\
\midrule
\ours
& 76.6 & 79.6
& 23.4 & 20.6
& 37.9 & 38.7 \\
\bottomrule
\end{tabular}
\caption{
Test-time scaling with Reflexion across one, two, and three sequential attempts.
\ours is included as a reference and is evaluated under the standard
single-attempt protocol.
}
\label{tab:reflexion_scaling}
\end{table}

\subsection{Internalization of Privileged Supervision}
\label{sec:internalization}

\begin{figure}[t]
    \centering
    \includegraphics[width=0.95\linewidth]{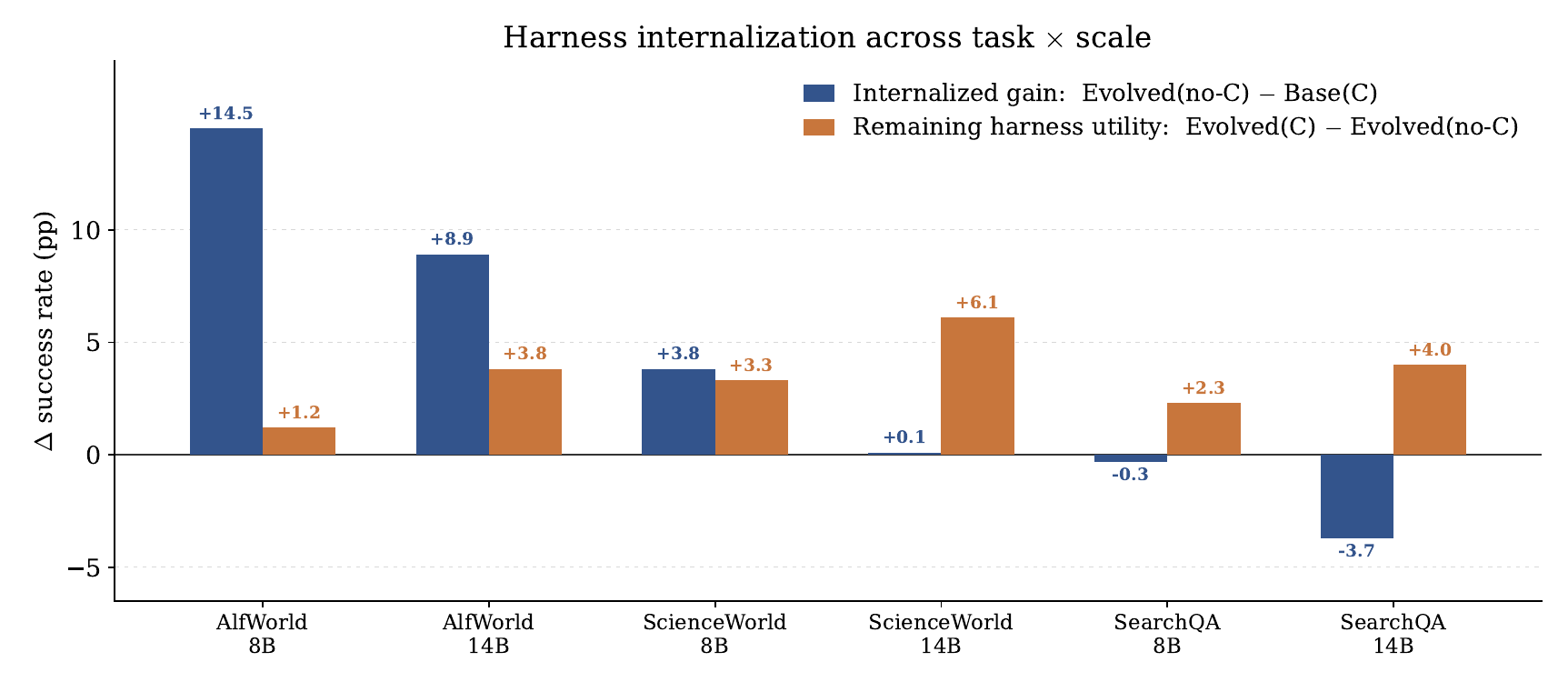}
    \caption{
        Internalization of privileged supervision across tasks and model scales.
        The blue bars show the internalized gain, measured as Evolved(no-C) $-$ Base(C), while the orange bars show the remaining harness utility, measured as Evolved(C) $-$ Evolved(no-C).
        Positive blue values indicate that the evolved model without privileged context outperforms the base model equipped with the harness at inference, whereas positive orange values indicate complementary information that remains useful after evolution.
    }
    \label{fig:harness_internalization}
\end{figure}

\paragraph{Internalization of Privileged Context.}
We next test whether the benefit of privileged supervision is transferred into the model itself rather than remaining dependent on inference-time scaffolding.
For each benchmark and model scale, we evaluate four conditions: Base(no-C), the pretrained base model without privileged context; Base(C), the same base model with privileged context supplied at inference time; Evolved(no-C), the self-evolved model evaluated without privileged context; and Evolved(C), the evolved model additionally provided with privileged context at inference time.
The key internalization comparison is Evolved(no-C) versus Base(C): if the evolved model without privileged context outperforms the base model with privileged context, then useful information from privileged training has been transferred into the model itself rather than remaining dependent on inference-time context.
As shown in \figref{fig:harness_internalization}, ALFWorld exhibits strong internalization at both model scales: Evolved(no-C) exceeds Base(C) by 14.5 and 8.9 percentage points for 8B and 14B, respectively, while reintroducing the harness to the evolved model provides only a small additional gain.
ScienceWorld shows more partial and scale-dependent internalization.
At 8B, Evolved(no-C) exceeds Base(C) by 3.8 points, while Evolved(C) provides a further 3.3-point improvement; at 14B, Evolved(no-C) is nearly identical to Base(C), whereas Evolved(C) adds 6.1 points, indicating that much of the privileged information remains complementary rather than being absorbed into the model.
SearchQA shows no measurable internalization, with Evolved(no-C) failing to outperform Base(C), consistent with the limited gains observed across self-evolution rounds.
Together, these results show that self-evolution can convert privileged retrospective supervision into persistent model capability, but the extent of internalization varies across tasks and scales.

\paragraph{Uptake of Retrospective Corrections.}
To more directly test whether privileged supervision affects the subsequent policy, we perform an offline re-scoring analysis on preserved consecutive checkpoints.
Specifically, for trajectories generated at round $r$, we keep the realized action tokens fixed and evaluate their probabilities under both the current policy $\pi^{(r)}$ and the subsequent policy $\pi^{(r+1)}$, while using the privileged teacher distribution to identify actions that the teacher disfavors.
If a realized action is assigned lower probability by $\pi^{(r+1)}$ than by $\pi^{(r)}$, this indicates that the next policy has moved away from a direction discouraged by the privileged teacher.
As shown in \tabref{tab:correction_uptake}, this directional correction appears in three of four tested round pairs, with the strongest effects early in evolution: $\Delta\log p=-0.384$ for ALFWorld $r1\!\rightarrow\!r2$ and $-0.696$ for ScienceWorld $r1\!\rightarrow\!r2$.
The effect becomes substantially weaker in later rounds.
Importantly, an exact-context re-score using teacher distributions logged during actual training does not support the stronger hypothesis that larger teacher corrections lead to proportionally larger uptake.
We therefore interpret these results as evidence that the \emph{direction} of privileged correction can be transferred into subsequent policies, rather than that the student simply imitates the full privileged teacher distribution.
Additional details of the re-scoring protocol and metrics are provided in \appref{sec:correction_uptake}.

\begin{table}[t]
\centering
\small
\setlength{\tabcolsep}{8pt}
\begin{tabular}{lc}
\toprule
\textbf{Round Pair} & \textbf{$\Delta \log p$ on Disfavored Actions} \\
\midrule
ALFWorld $r1\!\rightarrow\!r2$     & $-0.384$ \\
ALFWorld $r2\!\rightarrow\!r3$     & $-0.063$ \\
ScienceWorld $r1\!\rightarrow\!r2$ & $-0.696$ \\
ScienceWorld $r5\!\rightarrow\!r6$ & $+0.006$ \\
\bottomrule
\end{tabular}
\caption{
Directional uptake of privileged corrections across consecutive evolution rounds.
Negative values indicate that the subsequent policy assigns lower probability to realized actions disfavored by the privileged teacher.
Detailed analyses are provided in \appref{sec:correction_uptake}.
}
\label{tab:correction_uptake}
\end{table}

\subsection{Self-Evolution Across Rounds.}
\label{sec:self_evolution_analysis}

\begin{figure}[t]
    \centering
    \begin{subfigure}[t]{0.32\textwidth}
        \centering
        \includegraphics[width=\linewidth]{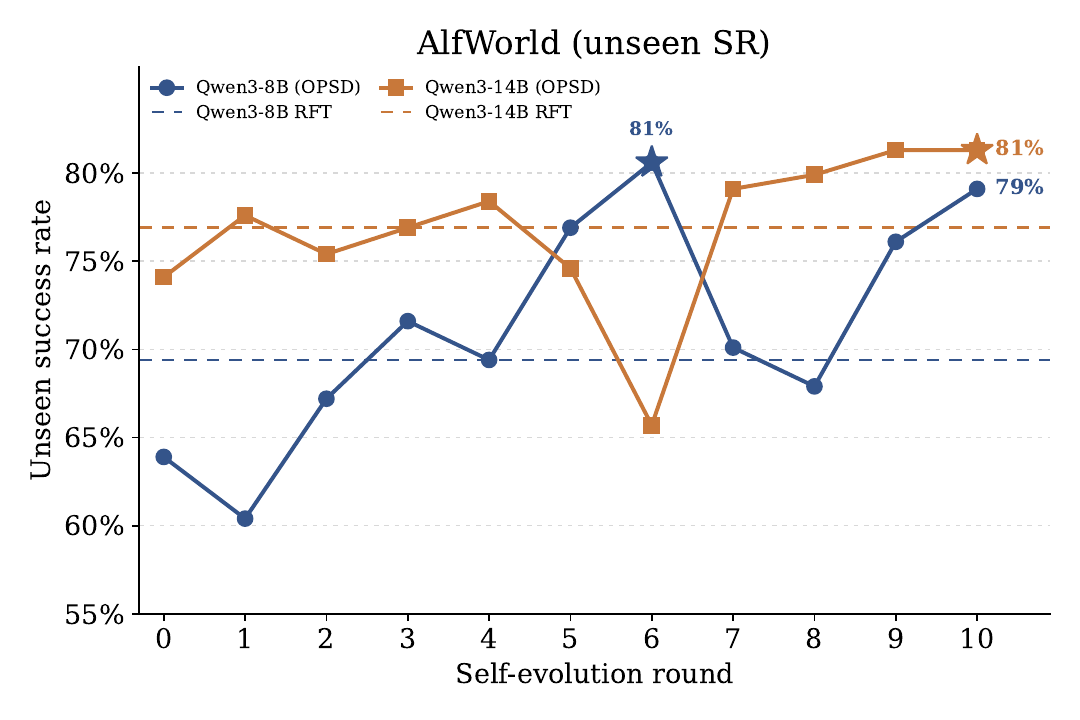}
        \caption{ALFWorld}
        \label{fig:evolution_alfworld}
    \end{subfigure}
    \hfill
    \begin{subfigure}[t]{0.32\textwidth}
        \centering
        \includegraphics[width=\linewidth]{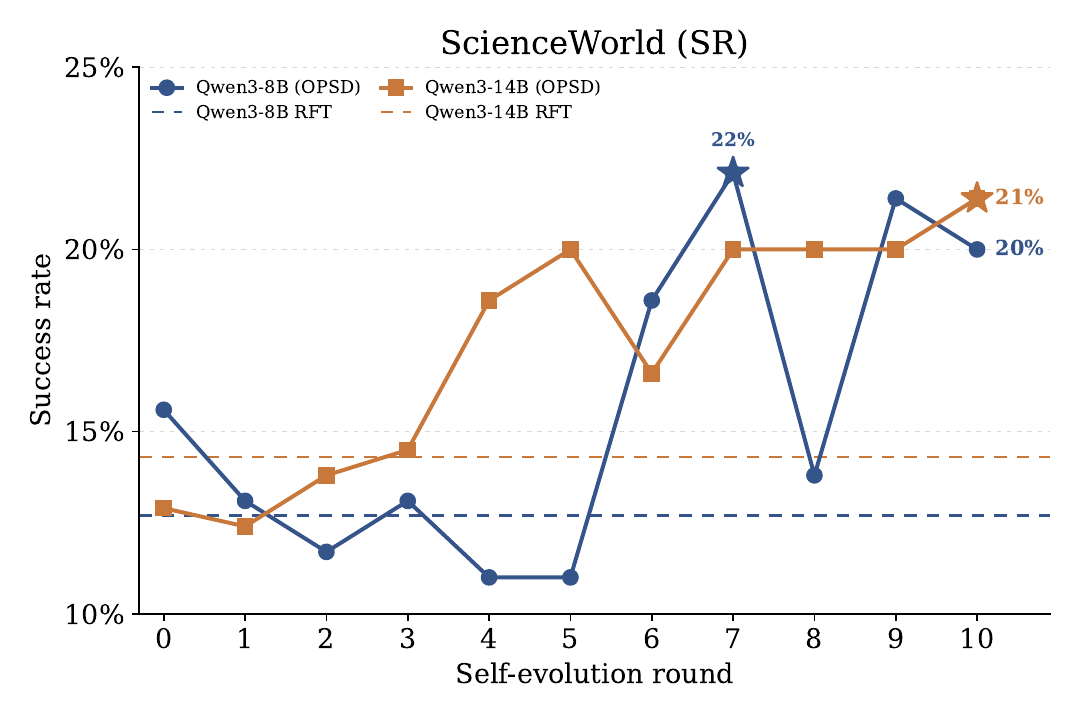}
        \caption{ScienceWorld}
        \label{fig:evolution_scienceworld}
    \end{subfigure}
    \hfill
    \begin{subfigure}[t]{0.32\textwidth}
        \centering
        \includegraphics[width=\linewidth]{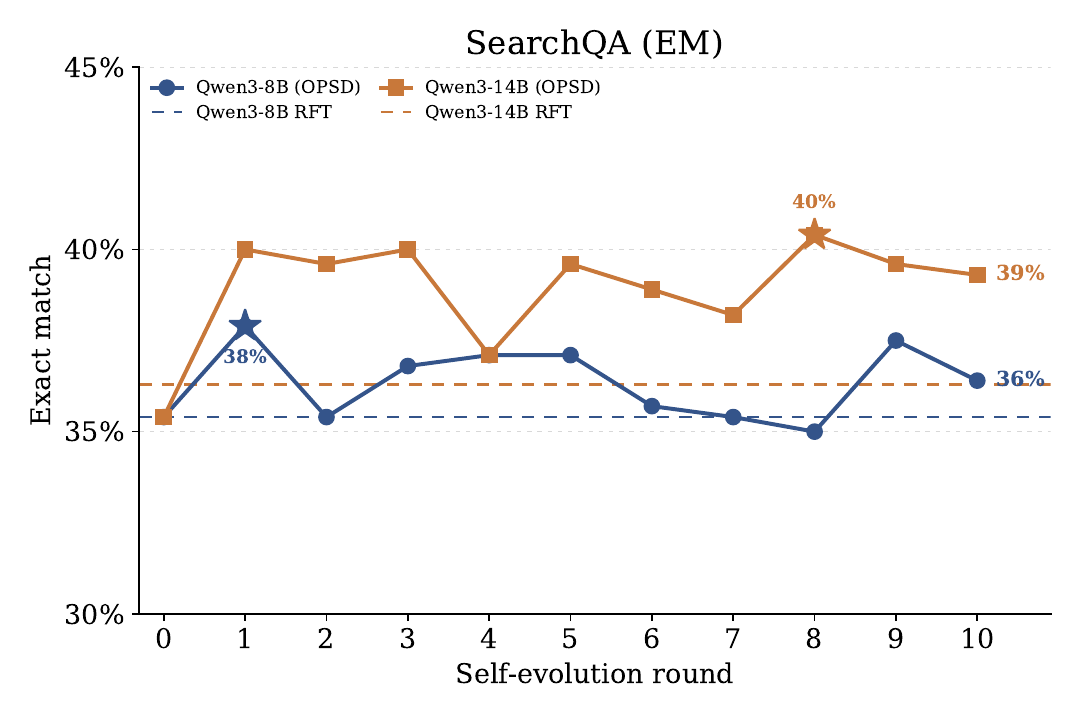}
        \caption{SearchQA}
        \label{fig:evolution_searchqa}
    \end{subfigure}

    \caption{
    Self-evolution trajectories of Qwen3-8B and Qwen3-14B over ten rounds.
    Round 0 denotes the corresponding base model, and dashed lines indicate the RFT baseline.
    Each panel shows a single-seed trajectory; stars denote the best round within the displayed run.
    ALFWorld reports unseen SR, while the main table reports the mean over seen and unseen splits. Canonical larger-scale evaluations are reported in \tabref{tab:main_results}.
    }
    \label{fig:self_evolution_curves}
\end{figure}

\figref{fig:self_evolution_curves} shows representative self-evolution trajectories over ten rounds for both model scales.
On ALFWorld, both Qwen3-8B and Qwen3-14B progressively reach substantially higher success rates than their initial models and the RFT baselines, although improvement is not strictly monotonic across rounds.
A similar pattern emerges on ScienceWorld: performance initially fluctuates around the baseline but improves noticeably in later rounds, indicating that useful behaviors can accumulate gradually through repeated self-evolution.
In contrast, SearchQA remains within a relatively narrow performance band throughout training, consistent with the limited gains observed in Table~\ref{tab:main_results}.
Additional teacher re-scoring diagnostics show a similarly stable correction signal (\appref{sec:teacher_diagnostics}).
Together, these trajectories show that repeated self-evolution can progressively improve interactive agent behavior on ALFWorld and ScienceWorld, while also revealing that the effectiveness of self-evolution depends on whether the task provides learnable behavioral headroom.

\subsection{Lesson-Source Ablation}
\label{sec:lesson_source_ablation}

Our default lesson construction summarizes both successful and failed trajectories
from accumulated past experience.
To examine whether failed experience provides useful information independently,
we restrict the lesson source to either successful trajectories only or failed
trajectories only, while keeping all other components fixed.
We evaluate Qwen3-8B for ten self-evolution rounds.

\begin{table}[t]
\centering
\caption{
Ablation on the source of accumulated lessons.
We report success rate (\%) with Qwen3-8B.
The default configuration summarizes both successful and failed trajectories and reports the main-result reference.
}
\label{tab:lesson_source_ablation}
\begin{tabular}{lcc}
\toprule
\textbf{Lesson Source} & \textbf{ALFWorld} & \textbf{ScienceWorld} \\
\midrule
Successful only & 69.4 & 22.5 \\
Failed only     & 72.6 & 22.5 \\
Successful + Failed (default) & 76.6 & 23.4 \\
\bottomrule
\end{tabular}
\end{table}

As shown in \tabref{tab:lesson_source_ablation}, the default configuration that summarizes both successful and failed trajectories achieves the best performance on both environments.
Restricting lessons to either successful or failed trajectories yields lower performance than the default main-result reference, while failed-only lessons remain competitive with successful-only lessons.
These results suggest that both successful and failed experience can provide useful retrospective information, although we treat this single-seed study as a diagnostic rather than a statistically matched comparison.

\subsection{Re-Scoring Analysis Details}
\label{sec:correction_uptake}

\paragraph{Protocol.}
We perform an offline re-scoring analysis on preserved consecutive evolution checkpoints to examine how retrospective teacher corrections are reflected in the subsequent policy.
For each round pair $r\!\rightarrow\!r+1$, we keep the round-$r$ trajectories and their realized tokens fixed and evaluate the same tokens under both the current policy $\pi^{(r)}$ and the subsequent policy $\pi^{(r+1)}$.
The privileged teacher distribution $q^{(r)}$ is used to identify whether each realized action is preferred or disfavored by the retrospective supervisor.
Because the trajectories are fixed, this analysis isolates changes in the policy distribution without introducing additional rollout variation or training.

\paragraph{Teacher Re-Scoring.}
We consider both proxy and privileged-context-conditioned teacher distributions.
The initial analysis uses the unconditioned base model as a proxy teacher, while the privileged-context analysis conditions the teacher on the round-specific retrospective context.
For ALFWorld, the corresponding precedent memory is reconstructed from the preserved round-level experience bank.
For the exact-context analysis, we additionally log the teacher distributions produced during training, allowing the same teacher targets to be evaluated directly without reconstructing the context.

\paragraph{Metrics.}
We measure uptake using complementary token-level statistics.
First, we compute the change in KL divergence between the teacher distribution and the policy across consecutive checkpoints.
Second, we measure the change in probability assigned to the teacher's top-1 token.
Finally, we measure the change in log probability of realized action tokens disfavored by the teacher, where a negative value indicates that the subsequent policy assigns lower probability to an action discouraged by privileged supervision.
Action tokens and ordinary language tokens are analyzed separately.

\paragraph{Evaluated Round Pairs.}
We analyze consecutive checkpoints from both ALFWorld and ScienceWorld, including early and later stages of self-evolution.
For the main directional-correction analysis, we evaluate ALFWorld $r1\!\rightarrow\!r2$ and $r2\!\rightarrow\!r3$, together with ScienceWorld $r1\!\rightarrow\!r2$ and $r5\!\rightarrow\!r6$.
The exact-context analysis additionally uses teacher distributions logged directly during training to avoid ambiguity introduced by context reconstruction.

\subsection{Teacher Re-Scoring Diagnostics}
\label{sec:teacher_diagnostics}

\paragraph{Task-Specific Correction Profiles.}
We additionally record per-turn teacher re-scoring statistics throughout self-evolution.
We define the correction gap as the difference between the log probability of the teacher's top-1 token and that assigned to the student's realized token; the correction rate as the fraction of tokens for which the teacher top-1 prediction differs from the student's token; teacher entropy as the average token-level entropy of the privileged teacher distribution over supervised action tokens; and the hit rate as the fraction of student tokens contained in the teacher's top-$k$ support used for distillation.
ALFWorld exhibits relatively small corrections throughout evolution, with correction gaps around $0.2$--$0.3$ and correction rates below approximately $7\%$.
ScienceWorld exhibits substantially larger corrections, with the 8B correction gap increasing from $0.809$ at round~1 to $1.420$ at round~7 and correction rates around $0.4$--$0.55$.
SearchQA instead shows moderate but comparatively stable correction gaps of approximately $0.5$ after the initial round.
Across all three benchmarks, the teacher top-$k$ hit rate remains approximately $0.97$--$1.00$, indicating that the truncated distillation targets retain high coverage of the student's realized tokens.
These task-specific correction profiles are consistent with the distinct evolution dynamics observed in the main experiments, although the correction gap alone does not establish whether the subsequent policy absorbs the corresponding supervision.


\subsection{Detailed Context-Evolution Analysis}
\label{sec:context_evolution_diagnostic}

\begin{wrapfigure}{r}{0.45\textwidth}
    \centering
    \includegraphics[width=\linewidth]{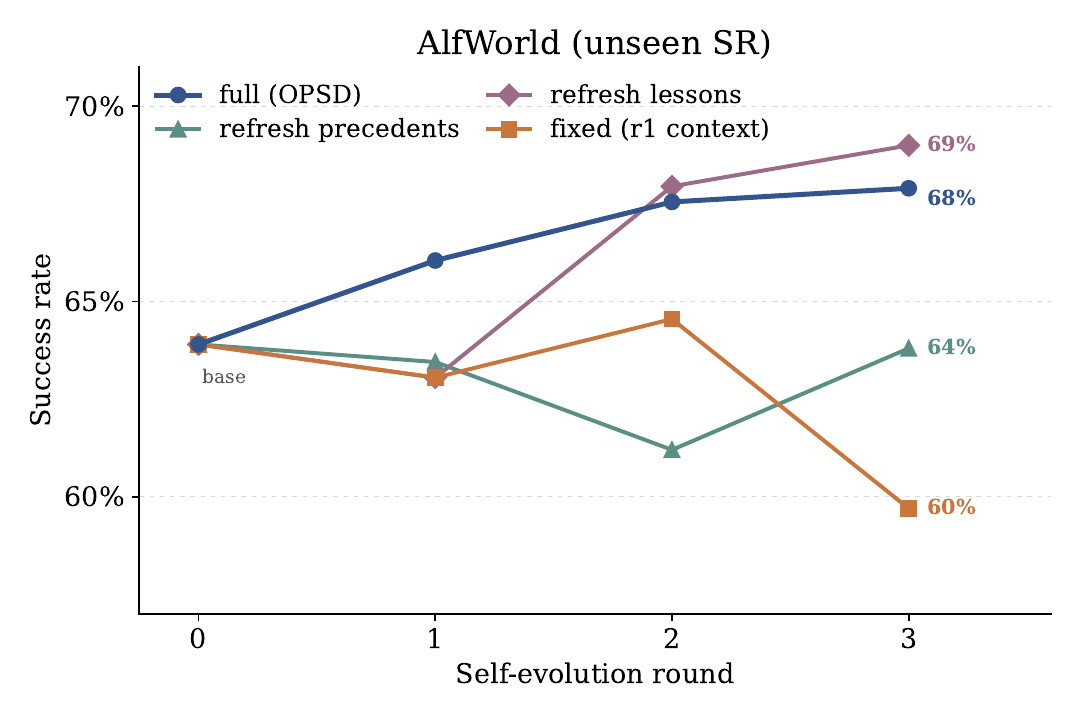}
    \vspace{-0.3in}
    \caption{
    Context evolution on ALFWorld-8B.
    Refreshing privileged context across self-evolution rounds substantially outperforms freezing the round-1 context, with lesson updates providing particularly strong gains.
    }
    \label{fig:context_evolution}
    \vspace{-0.05in}
\end{wrapfigure}

We next examine whether evolving the privileged context with newly collected experience is itself important for continued self-evolution.
Starting from the same base checkpoint, we train each variant for three self-evolution rounds and compare the standard evolving-context configuration of \ours against variants that evolve only lessons, evolve only precedents, or freeze the round-1 context for all subsequent rounds.
As shown in \figref{fig:context_evolution}, all variants begin from the
same performance at round 0, while their trajectories increasingly diverge as different context updates accumulate.
By round 3, the full evolving-context configuration reaches $67.9\%$ unseen SR, while refreshing lessons alone reaches $69.0\%$ and refreshing precedents alone reaches $63.8\%$; in contrast, freezing the round-1 context drops to $59.7\%$.
This suggests that, on ALFWorld, refreshing generated retrospective information is particularly important, consistent with the component ablation in \tabref{tab:context_ablation}.
Overall, these results show that repeatedly training against stale retrospective context limits continued improvement, whereas refreshing the context with newly collected experience supports continued self-evolution.
A diagnostic precedent-count study further reveals task-dependent context sensitivity (\appref{sec:precedent_count_sensitivity}), motivating an adaptive context policy rather than a single fixed configuration.
This task-dependent sensitivity further motivates the adaptive context-policy extension studied in \secref{sec:frontier_method}.

\subsection{Frontier Probe Construction and Validation}
\label{sec:frontier_probe}

\paragraph{No-Training Context Probe.}
To provide the frontier model with an inexpensive empirical signal for ranking candidate context-policy edits, we use a no-training privileged-context-at-inference probe.
For each candidate configuration, we render the corresponding privileged context using the same context-construction pipeline used during training, inject it into the model at inference time, and evaluate task success on a small training-distribution slice.
The resulting success rate is used as a proxy for the utility of the candidate context policy, requiring only inference and no parameter updates.
To reduce ranking noise, probe evaluation uses greedy decoding ($T=0$).

\paragraph{Validation on ScienceWorld.}
We validate the probe on ScienceWorld, where the controlled context ablation indicates substantial headroom from privileged-context construction.
In particular, removing precedents decreases dev SR from $23.4$ to $15.1$, providing a known context-quality ordering against which the probe can be evaluated.
Using four representative context configurations, the no-training probe produces SRs of $25.6$, $23.1$, $18.0$, and $13.3$ for the full, no-lessons, base, and no-precedents configurations, respectively.
The corresponding full-training evaluation obtains $23.4$, $22.7$, $15.6$, and $15.1$.
Thus, the probe recovers the same ordering across these configurations, with Spearman correlation $1.000$ and Pearson correlation $0.943$.
This provides evidence that the no-training probe supplies a useful ranking signal for context-policy search in a setting with meaningful context headroom.

\paragraph{Deterministic Evaluation.}
We found that stochastic decoding can introduce substantial variance into the probe score and obscure relatively small differences between candidate configurations.
We therefore use greedy decoding for all frontier candidate probes, while retaining the standard benchmark decoding configuration for the corresponding full-round evaluations.
For example, on ScienceWorld the greedy precedent-count probe yields SRs of $19.2$, $22.9$, $20.9$, and $20.0$ for counts $0$--$3$, respectively, correctly identifying that introducing precedents is beneficial while also indicating that more context is not necessarily better.

\paragraph{Context Fidelity.}
Because the probe is intended to estimate the utility of the same privileged information used during distillation, we ensure that probe-time context rendering matches the training-time format.
We audit the context-construction paths and verify byte-identical rendered notes across tested tasks, precedent counts, and lesson settings after resolving an earlier formatting mismatch.
This ensures that the probe evaluates the actual privileged-context representation used by the teacher rather than an independently reconstructed approximation.

\paragraph{Empirical Feedback Enables Frontier Optimization.}
\begin{wraptable}{r}{0.45\textwidth}
    \centering
    \small
    \setlength{\tabcolsep}{5pt}
    \begin{tabular}{lccc}
    \toprule
    \textbf{Method} & \textbf{Probe} & \textbf{SR} \\
    \midrule
    Fixed                 & --  & 23.4 \\
    Random                & --  & 23.2 \\
    Frontier (diagnostics) & No  & 23.9 \\
    Frontier + Probe      & Yes & \textbf{27.5} \\
    \bottomrule
    \end{tabular}
    \caption{
        Probe ablation on ScienceWorld-8B.
        The diagnostics-only frontier receives the same training digest but does not observe the no-training candidate probe.
    }
    \label{tab:frontier_probe_ablation}
\end{wraptable}

We next ablate the no-training probe used to evaluate candidate context-policy edits.
As shown in \tabref{tab:frontier_probe_ablation}, removing the probe reduces the ScienceWorld-8B SR from 27.5 to 23.9, placing the diagnostics-only frontier, which relies only on previous-round checkpoint accuracy and training metrics, in approximately the same regime as the fixed (23.4) and random-search (23.2) controls.
Thus, access to a frontier model alone is insufficient to reliably optimize privileged-context construction; the empirical probe provides a task-grounded signal that allows the frontier model to distinguish promising context edits from plausible but ineffective alternatives.

\subsection{Discovered Context Policies and Frontier Evolution Case Study}
\label{sec:frontier_policy_analysis}

\paragraph{Frontier Models Discover Task-Relevant Context Policies.}

\begin{wraptable}{r}{0.45\textwidth}
    \centering
    \small
    \setlength{\tabcolsep}{1pt}
    \begin{tabular}{lcc}
    \toprule
    \textbf{Context Edit} & \textbf{Adopted@Peak} & \textbf{Ablation $\Delta$SR} \\
    \midrule
    Precedents $\uparrow$ & \textbf{4/4} & $-8.3$ \\
    Lessons off           & 3/4          & $-0.7$ \\
    Future off            & 1/4          & $-5.8$ \\
    \bottomrule
    \end{tabular}
    \caption{
        Context-policy edits discovered by the frontier model on ScienceWorld-8B.
        Adopted@Peak reports how many independent runs contain the edit in their peak-performing configuration.
        Ablation $\Delta$SR denotes the performance drop when the corresponding component is removed from the canonical context.
    }
    \label{tab:frontier_discovered_policy}
\end{wraptable}

We finally examine which context-policy edits are retained by the frontier-guided search.
As shown in \tabref{tab:frontier_discovered_policy}, all four independent ScienceWorld-8B runs increase the number of retrieved precedents in their peak-performing configurations, closely matching the controlled ablation in \tabref{tab:context_ablation}, where removing precedents causes the largest performance drop.
The frontier also prunes lessons in three runs and future hindsight in one run.
This does not contradict their positive contribution in the component ablation: the ablation measures their utility under the canonical context, whereas frontier evolution makes stage-dependent decisions after the model has already undergone several rounds of training and may have internalized part of the previously supplied information.
Thus, the value of a context component can change over the course of self-evolution, and the frontier adapts the context policy accordingly rather than following a single fixed template.
A representative round-by-round trajectory illustrating this adaptive context evolution is provided below.

\paragraph{Case Study of Frontier-Guided Context-Policy Evolution}

\begin{table}[t]
\centering
\small
\setlength{\tabcolsep}{5pt}
\begin{tabular}{cllcc}
\toprule
\textbf{Round} & \textbf{Context Before} & \textbf{Proposed Change}
& \textbf{SR} & \textbf{Decision} \\
\midrule
$r_2$
& Prec.=0, Lessons=on, Future=on
& Prec. $0\!\rightarrow\!2$
& 0.180 & Revert \\

$r_3$
& Prec.=0, Lessons=on, Future=on
& Prec. $0\!\rightarrow\!1$
& 0.258 & Keep \\

$r_4$
& Prec.=1, Lessons=on, Future=on
& Future $\text{on}\!\rightarrow\!\text{off}$
& 0.221 & Revert \\

$r_5$
& Prec.=1, Lessons=on, Future=on
& No change
& 0.216 & Hold \\

$r_6$
& Prec.=1, Lessons=on, Future=on
& Prec. $1\!\rightarrow\!2$
& 0.243 & Revert \\

$r_7$
& Prec.=1, Lessons=on, Future=on
& No change
& 0.227 & Hold \\

$r_8$
& Prec.=1, Lessons=on, Future=on
& Lessons off, Future off
& 0.263 & Keep \\

$r_9$
& Prec.=1, Lessons=off, Future=off
& Prec. $1\!\rightarrow\!2$
& \textbf{0.285} & Keep \\

$r_{10}$
& Prec.=2, Lessons=off, Future=off
& Retrieval rel.\,$\rightarrow$\,rand.
& 0.225 & Revert \\
\bottomrule
\end{tabular}

\caption{
Representative trajectory of frontier-guided privileged-context evolution on ScienceWorld (Qwen3-8B).
``Context Before'' denotes the last accepted configuration.
The frontier may either propose a constrained context-policy edit or retain the current configuration when no candidate appears beneficial.
SR reports the resulting checkpoint performance at each round; proposed edits are subsequently kept or reverted according to their post-training performance.
}
\label{tab:frontier_case_study}
\end{table}

Table~\ref{tab:frontier_case_study} illustrates a representative ScienceWorld trajectory.
The frontier first explores adding precedents, reverting an overly aggressive increase to two precedents before retaining a smaller increase to one.
It subsequently tests modifications to future hindsight and precedent count, reverting changes that do not improve the resulting checkpoint, while leaving the context policy unchanged at rounds~5 and~7 when no candidate edit appears beneficial.
Later, it retains a configuration that removes lessons and future hindsight and then increases the precedent count to two, reaching the best SR of 28.5\% at round~9.
This trajectory illustrates that frontier-guided context evolution is not a sequence of unconditional mutations: the outer loop alternates between exploration, retention, reversion, and abstention as the underlying agent evolves.

\subsection{Teacher-Schedule Ablation}
\label{sec:teacher_schedule_ablation}

The default training configuration uses an adaptive freeze-to-peak teacher.
To isolate the role of this schedule, we compare two endpoint strategies under the same fixed canonical context policy using Qwen3-8B on ALFWorld dataset over ten self-evolution rounds.
In the \textit{frozen} variant, the teacher remains the initial base model throughout training.
In the \textit{fully evolving} variant, the teacher is updated to the current student after every round, with freeze-to-peak disabled.
These two configurations respectively isolate a completely stationary teacher and an unconstrained co-evolving self-teacher.

\begin{table*}[t]
\centering
\caption{
ALFWorld teacher-schedule ablation with Qwen3-8B under the fixed canonical context policy.
We report mean success rate (\%) over seen and unseen splits for each self-evolution round.
Results are intended as a controlled diagnostic of teacher dynamics.
}
\label{tab:teacher_schedule_ablation}
\resizebox{\textwidth}{!}{%
\begin{tabular}{lcccccccccccc}
\toprule
\textbf{Teacher Schedule}
& \textbf{R1}
& \textbf{R2}
& \textbf{R3}
& \textbf{R4}
& \textbf{R5}
& \textbf{R6}
& \textbf{R7}
& \textbf{R8}
& \textbf{R9}
& \textbf{R10}
& \textbf{Peak}
& \textbf{Final} \\
\midrule
Frozen
& 65.0
& \textbf{68.2}
& 63.5
& 65.7
& 63.5
& 65.7
& 62.4
& 67.5
& 67.1
& 67.9
& 68.2
& 67.9
\\
Fully evolving
& 64.6
& 71.6
& 78.5
& 79.2
& 78.1
& 79.6
& \textbf{80.3}
& 76.2
& 39.4
& 0.0
& 80.3
& 0.0
\\
\bottomrule
\end{tabular}%
}
\end{table*}

As shown in \tabref{tab:teacher_schedule_ablation}, the frozen teacher remains within a narrow performance range throughout all ten rounds, reaching a peak mean success rate of $68.2\%$ and ending at $67.9\%$.
In contrast, the fully evolving teacher benefits strongly from co-evolution during the early rounds, reaching $80.3\%$ at round 7, but subsequently becomes unstable and collapses from $76.2\%$ at round 8 to $39.4\%$ at round 9 and $0\%$ at round 10.
These two behaviors motivate the default freeze-to-peak strategy: allowing the teacher to track the student initially provides adaptive supervision, while freezing the best validation checkpoint once performance deteriorates prevents the teacher from following the student into late-stage collapse.
We view this experiment as a diagnostic of the teacher-update mechanism rather than a statistically powered benchmark comparison, since each endpoint schedule is evaluated with a single seed.

\subsection{Sensitivity to Precedent Count}
\label{sec:precedent_count_sensitivity}

We further examine whether a fixed precedent budget transfers uniformly across environments.
Holding all other components of the canonical context policy fixed, we vary only the number of retrieved precedents $n\in\{1,2,4\}$ using Qwen3-8B.
The default configuration uses $n=2$.
Results are shown in \tabref{tab:precedent_count_sensitivity}.

\begin{table}[t]
\centering
\caption{
Sensitivity to the number of retrieved precedents.
We report success rate (\%) for Qwen3-8B.
$n=2$ denotes the default configuration and reports the default main-result reference.
}
\label{tab:precedent_count_sensitivity}
\begin{tabular}{lcc}
\toprule
\textbf{Precedent Count} & \textbf{ALFWorld} & \textbf{ScienceWorld} \\
\midrule
$n=1$ & 75.6 & \textbf{24.9} \\
$n=2$ (default) & 76.6 & 23.4 \\
$n=4$ & \textbf{78.8} & 20.3 \\
\bottomrule
\end{tabular}
\end{table}

The effect of precedent count differs substantially across environments. On ALFWorld, the $n=4$ variant performs strongly, whereas ScienceWorld performs better with fewer precedents.
Thus, additional context is not uniformly beneficial, and the appropriate precedent budget is task dependent.
This sensitivity further motivates adapting the context-construction policy rather than relying on a single fixed configuration across environments.

\section{Prompt Templates}
\label{sec:prompt_templates}

We provide the prompt templates used in our experiments, including (i) agent rollout and evaluation prompts, (ii) privileged retrospective-context prompts, and (iii) frontier-guided context-policy evolution prompts.

\paragraph{Privileged Retrospective-Context Prompts.}
Tables~\ref{tab:prompt_alfworld}--\ref{tab:prompt_searchqa} show the
task-specific prompts used by the acting policy during rollout and evaluation.
The one-shot demonstrations are fixed across self-evolution rounds.

\begin{table*}[t]
\centering
\caption{Prompt template for ALFWorld. A one-shot demonstration is provided before the user turn.}
\label{tab:prompt_alfworld}
\resizebox{0.95\textwidth}{!}{%
\begin{tabular}{lp{0.78\textwidth}}
\toprule
\textbf{Field} & \textbf{Prompt Content} \\
\midrule
System prompt &
\begin{minipage}[t]{0.78\textwidth}
\ttfamily
You are an embodied agent solving a household task in a text-based simulator (AlfWorld). You receive an Observation describing what you can see and a list of Admissible commands you may execute. On every turn you must reply with exactly two lines: Thought: <one short sentence of reasoning> Action: <one command, copied verbatim from the Admissible list when possible> Plan briefly, then issue the next action. Only the line beginning with 'Action:' is used to step the environment. Do not add any other text after the Action line.
\end{minipage}
\\
\midrule
One-shot demonstration &
\begin{minipage}[t]{0.78\textwidth}
\ttfamily
A one-shot demonstration.
\end{minipage}
\\
\midrule
User turn &
\begin{minipage}[t]{0.78\textwidth}
\ttfamily
Task: \{task\_goal\}\\
Observation: \{scene\_description\}\\
Admissible: cmd1; cmd2; ...; cmd30 (+N more)
\end{minipage}
\\
\midrule
Assistant turn &
\begin{minipage}[t]{0.78\textwidth}
\ttfamily
Thought: <reasoning>\\
Action: <command>
\end{minipage}
\\
\bottomrule
\end{tabular}%
}
\end{table*}

\begin{table*}[t]
\centering
\caption{Prompt template for ScienceWorld. A single generic one-shot ReAct trajectory is provided before the user turn.}
\label{tab:prompt_scienceworld}
\resizebox{0.95\textwidth}{!}{%
\begin{tabular}{lp{0.78\textwidth}}
\toprule
\textbf{Field} & \textbf{Prompt Content} \\
\midrule
System prompt &
\begin{minipage}[t]{0.78\textwidth}
\ttfamily
You are an embodied agent solving a science task in the ScienceWorld text-based simulator. You receive an Observation describing what you can see and an Inventory listing what you carry. On every turn you must reply with exactly two lines: Thought: <one short sentence of reasoning> Action: <one valid action> Plan briefly, then issue the next action. Only the line beginning with 'Action:' is used to step the environment.
\end{minipage}
\\
\midrule
One-shot demonstration &
\begin{minipage}[t]{0.78\textwidth}
\ttfamily
A single generic ReAct trajectory.
\end{minipage}
\\
\midrule
User turn &
\begin{minipage}[t]{0.78\textwidth}
\ttfamily
Task: \{task\_description\}\\
Observation: \{observation\}\\
Inventory: \{inventory or "(empty)"\}
\end{minipage}
\\
\midrule
Assistant turn &
\begin{minipage}[t]{0.78\textwidth}
\ttfamily
Thought: <reasoning>\\
Action: <valid action>
\end{minipage}
\\
\bottomrule
\end{tabular}%
}
\end{table*}

\begin{table*}[t]
\centering
\caption{Prompt template for SearchQA. The agent repeatedly issues Wikipedia search queries and terminates with a concise final answer.}
\label{tab:prompt_searchqa}
\resizebox{0.95\textwidth}{!}{%
\begin{tabular}{lp{0.78\textwidth}}
\toprule
\textbf{Field} & \textbf{Prompt Content} \\
\midrule
System prompt &
\begin{minipage}[t]{0.78\textwidth}
\ttfamily
You are an expert open-domain question-answering agent. You answer questions by repeatedly issuing wikipedia search queries, reading the returned snippets, and finally producing a concise short answer.

Output protocol -- on every turn you must reply with exactly two blocks: <thought>... brief reasoning (1-2 sentences) ...</thought> <action>VERB[PAYLOAD]</action> where VERB[PAYLOAD] is exactly one of: search[your concise wikipedia query, 3-8 words] to fetch top-3 snippets answer[your final short answer] to terminate the episode. Use search[...] until you have enough evidence, then answer[...]. Final answers should be short and direct (a single entity, date, or short phrase). Do not output anything after the </action> tag.
\end{minipage}
\\
\midrule
One-shot demonstration &
\begin{minipage}[t]{0.78\textwidth}
\ttfamily
A fixed one-shot demonstration.
\end{minipage}
\\
\midrule
User turn &
\begin{minipage}[t]{0.78\textwidth}
\ttfamily
Question: \{question\}\\
<information>1) ... 2) ... 3) ...</information>
\end{minipage}
\\
\midrule
Assistant turn &
\begin{minipage}[t]{0.78\textwidth}
\ttfamily
<thought>...</thought>\\
<action>search[query]</action>\\
\textit{or}\\
<thought>...</thought>\\
<action>answer[final]</action>
\end{minipage}
\\
\bottomrule
\end{tabular}%
}
\end{table*}

The interaction prompts above are used by the acting policy during both rollout and standard evaluation.
The privileged retrospective information described below is constructed only after a trajectory has been collected and is provided to the teacher during retrospective re-scoring; it is not available to the acting policy during standard rollout or evaluation.

\paragraph{Frontier-Guided Context-Policy Evolution Prompts.}
\begin{table*}[t]
\centering
\caption{Prompts used to generate privileged retrospective information.
The lessons summary is generated once per task type and round from selected
successful and failed trajectories, while reflection is generated retrospectively
for the current completed trajectory.}
\label{tab:privileged_generation_prompts}
\small
\begin{tabular}{p{0.16\textwidth}p{0.78\textwidth}}
\toprule
\textbf{Component} & \textbf{Prompt} \\
\midrule

Lessons &
\begin{minipage}[t]{0.78\textwidth}
\ttfamily
Successful and failed attempts at '\{task\_type\}' ALFWorld tasks:\\
\{demos\}\\[2pt]
In 3 short bullet lessons, give the GENERAL strategy for this task type: the correct
sub-goal order, and the common mistakes seen in the FAILED attempts to AVOID.
\end{minipage}
\\
\midrule

Reflection &
\begin{minipage}[t]{0.78\textwidth}
\ttfamily
You attempted an ALFWorld task.\\
TASK: \{goal\}\\[2pt]
YOUR ATTEMPT (step: action -> observation):\\
\{ftxt\}\\
OUTCOME: \{environment outcome\} after \{N\} steps.\\[2pt]
Using hindsight, identify the SINGLE earliest step whose action was the pivotal
mistake, and the better action there. Respond EXACTLY as:\\
STEP: <int>\\
ACTION: <action>
\end{minipage}
\\

\bottomrule
\end{tabular}
\end{table*}

\begin{table*}[t]
\centering
\caption{Construction of the privileged note used for teacher re-scoring.
Components are enabled or disabled by the corresponding context-policy
configuration. The default configuration uses lessons, two precedents,
reflection, and future hindsight.}
\label{tab:privileged_note_construction}
\small
\begin{tabular}{p{0.18\textwidth}p{0.76\textwidth}}
\toprule
\textbf{Stage} & \textbf{Construction Template} \\
\midrule

Forward context &
\begin{minipage}[t]{0.76\textwidth}
\ttfamily
[GENERAL STRATEGY across tasks] \{behavior\}\par
[LESSONS for this task type] \{lessons summary\}\par
[PRECEDENT 1 --- a different, similar task solved successfully] 1) act; 2) act; ...\par
[PRECEDENT 2 --- ...] 1) ...
\end{minipage}
\\
\midrule

Reflection &
\begin{minipage}[t]{0.76\textwidth}
\ttfamily
[SELF-REFLECTION on your current attempt] \{reflection\}
\end{minipage}
\\
\midrule

Future hindsight &
\begin{minipage}[t]{0.76\textwidth}
\ttfamily
[HINDSIGHT --- with knowledge of the future: ...]\\
\{future trajectory information\}
\end{minipage}
\\
\midrule

Final privileged note &
\begin{minipage}[t]{0.76\textwidth}
\ttfamily
[PRIVILEGED] [LESSONS for this task type] ...\\
\hspace*{1em}[PRECEDENT 1 ...] ...\\
\hspace*{1em}[PRECEDENT 2 ...] ...\\
\hspace*{1em}[SELF-REFLECTION on your current attempt] ...\\
\hspace*{1em}[HINDSIGHT --- with knowledge of the future: ...]
\end{minipage}
\\
\midrule

Default injection &
\begin{minipage}[t]{0.76\textwidth}
The privileged note is appended to the teacher's initial user turn
(\texttt{user\_end}). Alternative supported positions are
\texttt{system}, \texttt{user\_begin}, \texttt{last\_user}, and
\texttt{trailing}.
\end{minipage}
\\

\bottomrule
\end{tabular}
\end{table*}

\begin{table*}[t]
\centering
\caption{Frontier-model prompt for proposing candidate context-policy edits.
The frontier model receives round diagnostics, the current editable configuration,
and any no-training probe evidence accumulated during the current search round.}
\label{tab:frontier_hypothesis_prompt}
\small
\begin{tabular}{p{0.96\textwidth}}
\toprule
\textbf{Frontier hypothesis-generation prompt (P1)} \\
\midrule
\begin{minipage}[t]{0.96\textwidth}
\ttfamily
You are a SCIENTIST tuning the privileged context policy (the 'harness') for
on-policy self-distillation (OPSD). A teacher rescores the student's OWN rollouts
under a note built from precedents + lessons + behavior + future + reflection;
the harness config controls that note.\\
GOAL: maximize downstream dev SR (task success) of the next model.\\[4pt]

PRINCIPLES (general, NOT task-specific):\\
- MORE IS NOT BETTER. Extra/again-predicted context adds tokens+noise without
helping; 'remove/simplify' and turning a component OFF are as valid as 'add'.
EVERY whitelisted knob is fair game --- precedent count AND retrieval AND
lessons/behavior/future/reflection/selection/etc. Do NOT fixate on precedent count.\\
- You are optimizing PROXY DEV SR measured by a NO-TRAIN harness@inference eval:
each candidate's precedents are injected into the model's OWN context and we read
dev SR. This probe reflects the FORWARD-CONTEXT knobs (precedent count/retrieval)
--- it is VALIDATED to track true full-round SR for those. It does NOT move for
hindsight/training-mechanism knobs
(future/lessons/reflection/teacher\_temp/selection): those have no inference analog.
You MAY still propose them, but the probe will read $\sim$flat for them and they
will be judged only by the real training round (the retrospective gate), not this probe.\\
- Prefer edits your evidence says raise SR.\\
- Form hypotheses, test them, and revise.\\[4pt]

===== ROUND DIAGNOSTICS =====\\
\{digest\}\\
===== EDITABLE KNOBS (whitelist) =====\\
\{whitelist\}\\
===== PROBE EVIDENCE SO FAR THIS ROUND (harness@inference dev SR, no-train) =====\\
\{accumulated evidence, or ``(none yet --- this is the first iteration)''\}\\[4pt]

Propose \{NCAND\} DISTINCT candidate edits to TEST now
(each changes AT MOST \{MAX\_EDITS\} whitelisted knobs, in-bounds).
Cover DIFFERENT knob families (include >=1 subtractive/OFF and >=1 touching a
generated component: lessons/behavior/future/reflection).
Print EXACTLY \{NCAND\} lines, nothing else:\\
CAND: <key>=<val>[, <key2>=<val2>]; HYP: <why this should raise dev SR>
\end{minipage}
\\
\bottomrule
\end{tabular}
\end{table*}

\begin{table*}[t]
\centering
\caption{Frontier-model prompt for selecting among candidate context-policy edits
after no-training probe evaluation.}
\label{tab:frontier_decision_prompt}
\small
\begin{tabular}{p{0.96\textwidth}}
\toprule
\textbf{Frontier candidate-selection prompt (P2)} \\
\midrule
\begin{minipage}[t]{0.96\textwidth}
\ttfamily
Candidate edits tested this iteration
(harness@inference dev SR, no-train; higher is better):\\
\{cand1: <edit> -> proxy\_dev\_SR=...; cand2: ...; ...\}\\
CURRENT (unedited) harness under the SAME no-train probe:
proxy\_dev\_SR=\{CUR\_SR\}\\[4pt]

Read the proxy SRs as EVIDENCE of downstream task success. Choose the action:\\
- CHOOSE cand<i> --- if a candidate's proxy SR is convincingly the best
(and beats leaving the harness as-is).\\
- CHOOSE none --- if none convincingly beats the current harness.\\
- REHYPOTHESIZE --- if the evidence suggests a better untested direction;
you will get another probe round.\\[4pt]

Print EXACTLY one line: `CHOOSE: cand<i>' or `CHOOSE: none' or
`REHYPOTHESIZE'\\
Then one line: `WHY: <cite the proxy-SR evidence>'
\end{minipage}
\\
\bottomrule
\end{tabular}
\end{table*}

The frontier search alternates between candidate proposal and no-training
probe evaluation for at most two iterations per round.
The selected context-policy edit is then used for the subsequent full
self-evolution round and subjected to the retrospective acceptance criterion
in Eq.~\ref{eq:context_acceptance}.
We use an acceptance tolerance of $\epsilon=0.5\%$.

\end{document}